\documentclass[11pt]{article}
\usepackage{EACL2023}
\usepackage{amsmath}
\usepackage{times,latexsym}
\usepackage{url}
\usepackage[T1]{fontenc}
\usepackage{graphicx}
\usepackage{amssymb}
\usepackage{enumitem}
\usepackage{float}
\restylefloat{table}
\usepackage{placeins}
\usepackage{stfloats}
\usepackage{multirow}
\usepackage{microtype}
\usepackage{inconsolata}
\usepackage[export]{adjustbox}
\usepackage{inconsolata}
\usepackage{cleveref}
\newcommand{\travlr}{\textsc{TraVLR}}

\usepackage{subcaption}
\usepackage{xspace,mfirstuc,tabulary}

\usepackage{booktabs}

\newcommand{\tablesize}{\fontsize{6.8}{10}\selectfont}

\usepackage[normalem]{ulem}
\usepackage{xcolor, soul}
\definecolor{kjcolor}{rgb}{0.858, 0.188, 0.478}
\newcommand{\kj}[1]{#1}

\usepackage{soul}
\definecolor{red}{HTML}{fcdb8f}
\definecolor{green}{HTML}{D0ECA6}
\definecolor{pink}{HTML}{ffc1e7}

\DeclareRobustCommand{\hlred}[1]{{\sethlcolor{red}\hl{#1}}}

\DeclareRobustCommand{\hlgreen}[1]{{\sethlcolor{green}\hl{#1}}}

\title{\travlr: Now You See It, Now You Don't!\\ A Bimodal Dataset for Evaluating Visio-Linguistic Reasoning}

\author{
 Keng Ji Chow$^{\natural\lambda}$\thanks{\; Equal contribution} \qquad Samson Tan$^{\natural\S}$\footnotemark[1] \qquad Min-Yen Kan$^\natural$
 \\$^\natural$Department of Computer Science, National University of Singapore \\
 $^\lambda$Department of English Language and Literature, National University of Singapore
 \\$^\S$AWS AI Research \& Education \vspace{0.3em}\\
 kengjichow@u.nus.edu \\
 \{samson.tmr, kanmy\}@comp.nus.edu.sg
}

\begin{document}
\maketitle
\begin{abstract}
Numerous visio-linguistic (V+L) representation learning methods have been developed, yet existing datasets do not adequately evaluate the extent to which they represent visual and linguistic concepts in a unified space. We propose several novel evaluation settings for V+L models, including \emph{cross-modal transfer}. Furthermore, existing V+L benchmarks often report global accuracy scores on the entire dataset, making it difficult to pinpoint the specific reasoning tasks that models fail and succeed at. We present \travlr, a synthetic dataset comprising four V+L reasoning tasks. \travlr's synthetic nature allows us to constrain its training and testing distributions along task-relevant dimensions, enabling the evaluation of out-of-distribution generalisation. Each example in \travlr\ redundantly encodes the scene in two modalities, allowing either to be dropped or added during training or testing without losing relevant information. We compare the performance of four state-of-the-art V+L models, finding that while they perform well on test examples from the same modality, they all fail at cross-modal transfer and have limited success accommodating the addition or deletion of one modality. We release \travlr\ as an open challenge for the research community.\footnote{Code and dataset available at \url{https://github.com/kengjichow/TraVLR}.}
\end{abstract}

\section{Introduction}

Research in psycholinguistics has found that human processing of spatial words activates brain regions associated with the visual system \citep{tang2021cortical}, suggesting the latter's involvement in processing linguistic input. It is reasonable to expect multimodal neural models to \kj{resemble humans in being able to leverage capabilities in the visual domain to solving problems in the text domain, and vice versa}. Following its recent success in the text domain \citep{devlin2019bert}, the pretraining--fine-tuning paradigm has been applied to the vision and text modalities to create unified visio-linguistic (V+L) representations. Just as pretrained multilingual models have been shown capable of zero-shot cross-lingual transfer on various NLP tasks \citep{conneau2020unsupervised}, we may expect \emph{true} V+L models to be capable of generalising to a modality not seen during fine-tuning. 

\begin{figure}[t!]
\small
\centering
\begin{subfigure}{0.45\textwidth}
\includegraphics[width=\textwidth]{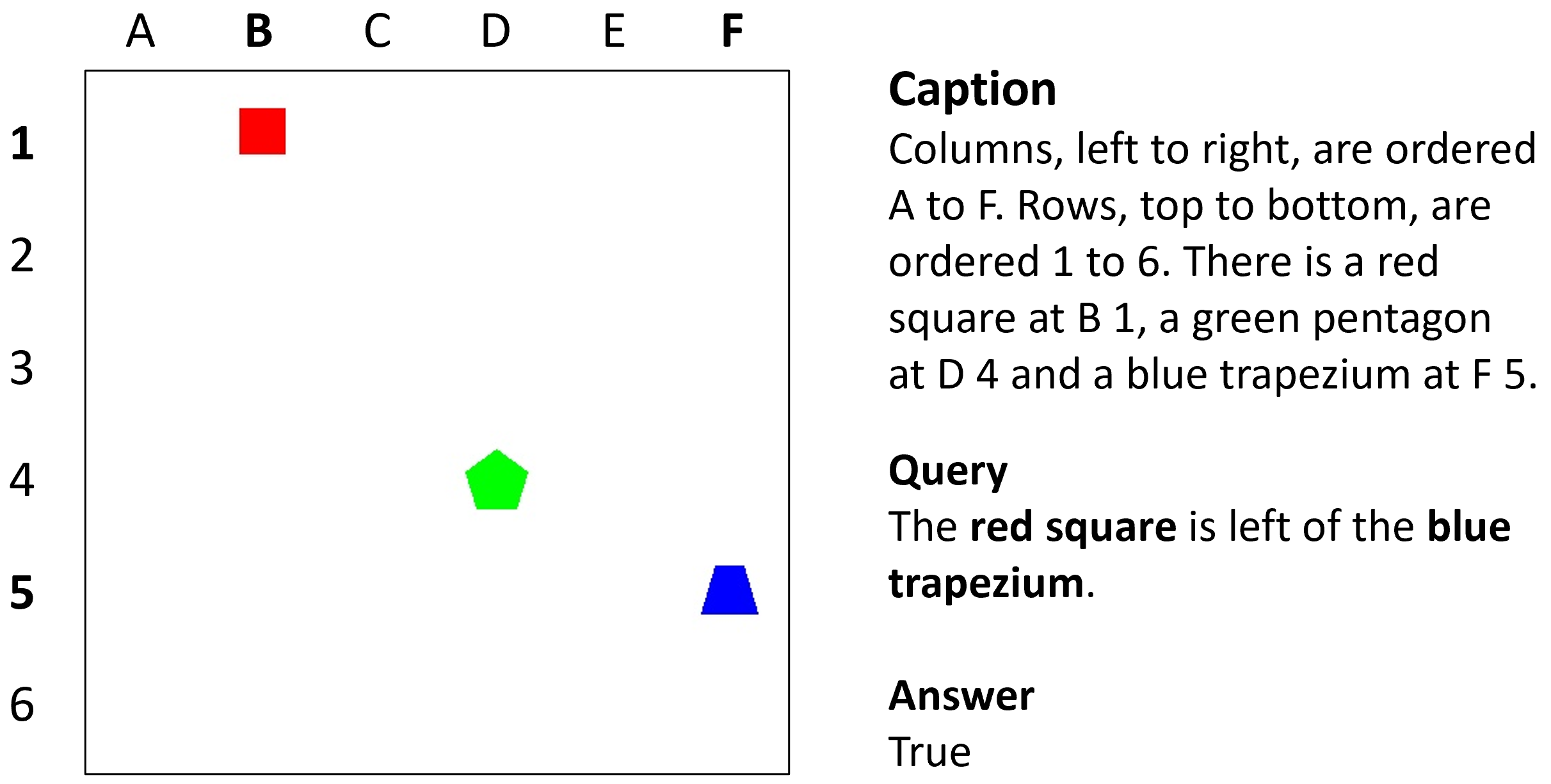}
\caption{A complete example for the spatiality task.}
\end{subfigure}

\begin{subfigure}{0.45\textwidth}
\vspace{0.35em}
\includegraphics[width=\textwidth]{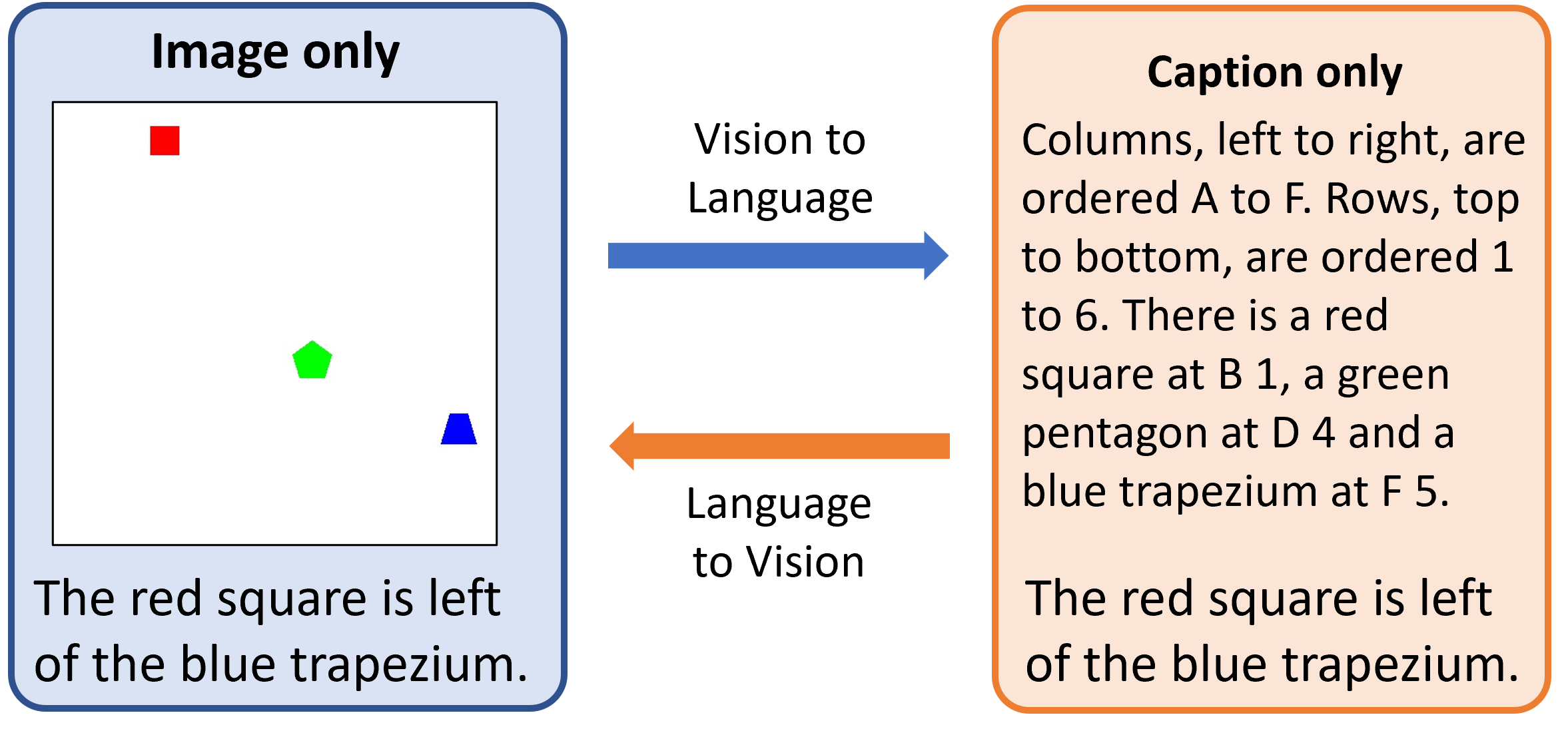}
\caption{Possible directions of cross-modal transfer.}
\end{subfigure}
\caption{An example from \travlr\ (a). Both image and caption fully represent the scene; either can be dropped during training/testing, enabling the evaluation of cross-modal transfer ability (b).}
\label{fig:intro_travlr}
\vspace{-4mm}
\end{figure}

However, current approaches for benchmarking V+L models involve reporting global accuracy scores on the entire dataset, rendering the specific sources of success and failure difficult to diagnose
\citep{ribeiro2020beyond, goel2021robustness}. For instance, Visual Question Answering (VQA, \citealt{goyal2017making}) tasks may allow models to exploit dataset bias \citep{dancette2021beyond}, or may reduce to object recognition problems which do not evaluate the models' ability to perform more complex tasks, beyond aligning words or phrases in the text to a portion of the image \citep{hudson2019gqa, acharya2019tallyqa}. As \citet{bernardi2021linguistic} note, the ability to reason over multiple objects and answer relational questions is crucial to the genuine mastery of language.

Datasets such as NLVR2 \citep{suhr2019corpus} address this limitation, but do not allow for fine-grained evaluation along specific dimensions \citep{tan-etal-2021-reliability}. CLEVR \citep{johnson2017clevr} and \textsc{ShapeWorld} \citep{kuhnle2017shapeworld} enable targeted evaluations of a V+L model's reasoning abilities but only encode the scene unimodally, as images. Additionally, their test examples may still be in the training distribution with respect to task-relevant dimensions, making it difficult to draw conclusions about generalisation ability.

We thus contribute \travlr, a synthetic dataset comprising four V+L reasoning tasks: spatiality, cardinality, quantifiers, and numerical comparison, all of which require reasoning over multiple objects and have been shown to be challenging for V+L models \citep{johnson2017clevr, parcalabescu2020seeing}. Unlike \textsc{ShapeWorld}, we fully leverage the benefits of using a synthetic dataset by controlling the train-test split such that examples in the out-of-distribution (OOD) test set are OOD \emph{with respect to task-relevant dimensions}. We thus argue that \travlr\ serves as a basic sanity check for the abstract reasoning and out-of-distribution generalisation capabilities of models, and is complementary to datasets that evaluate real-world object recognition and compositional reasoning abilities, such as GQA \citep{hudson2019gqa}. 

Inspired by the word/picture sentence verification task from psycholinguistics \citep{goolkasian1996picture}, we further propose various novel evaluation settings by representing the scene bimodally as both an image and a caption. First, \travlr\ supports the novel \emph{cross-modal transfer} setting (\Cref{fig:intro_travlr}): If pretrained V+L models have learnt a truly multimodal representation, they should be able to learn a reasoning task with input from one modality and perform inference using input from the other modality with little to no extra training. Being able to transfer cross-modally in a zero- or few-shot manner may improve data efficiency in applications where diverse image data is difficult to obtain. Furthermore, models should also succeed on test settings where either an unseen modality is added, or a seen modality is dropped. 

Using \travlr, we perform extensive analysis of the ability of four Transformer-based V+L models to perform various reasoning tasks. We show that current V+L models:

\vspace{-0.4em}
\begin{itemize}[leftmargin=1em]
\itemsep-0.3em 
    \item May require unreasonably large amounts of data to learn simple visio-linguistic reasoning tasks.
    \item Exhibit a strong textual bias.
    \item Are unable to perform cross-modal transfer. We thus pose this as an open challenge for future V+L models.
\end{itemize}

\section{Related Work}

\paragraph{V+L tasks and datasets.}

The Visual Question Answering (VQA) task involves answering a question about an image. It is a complex task as it requires an ability to process input in both visual and textual modalities \citep{antol2015vqa}. A known issue with VQA datasets is the presence of real-world language priors and statistical biases in the training and testing distribution \citep{kervadec2021roses, agrawal2018don, kafle2019challenges}. Although VQA~v2.0 \cite{goyal2017making} was improved by balancing each query with pairs of images, \citet{dancette2021beyond} show that it still contains both unimodal and multimodal biases that models can exploit. Furthermore, questions in VQA may use non-compositional language that does not require abilities beyond object recognition. 

NLVR \citep{suhr2017corpus} addresses the lack of compositionality in VQA using synthetic images of abstract 2D shapes, accompanied by human-written English sentences to be judged true or false. NLVR2 \citep{suhr2019corpus} and SNLI-VE \citep{xie2019visual} also involve truth-value/entailment judgement tasks, but use photographs instead of synthetic images. Both lack detailed annotations of the specific semantic phenomena evaluated by each example. GQA improves over VQA by focusing on compositional questions that require reasoning over multiple objects and contains detailed annotations \citep{hudson2019gqa}, but still suffers from statistical imbalances and the lack of an out-of-distribution test set \citep{kervadec2021roses}.

Other synthetic datasets focusing on reasoning include CLEVR \citep{johnson2017clevr}, a fully synthetic and annotated 3D dataset, and \textsc{ShapeWorld}, a 2D dataset targeting linguistic phenomena such as spatial relationships and quantifiers. SPARTQA \citep{mirzaee2021spartqa} is another 2D synthetic dataset built upon NLVR focusing on spatial reasoning among other linguistic phenomena. gSCAN \citep{ruis2020benchmark} focuses on generalisation of commands within a 2D grid-world. 

\paragraph{V+L models.}

Pretrained V+L models differ in their architecture and pretraining methods. VL-BERT \citep{su2019vl}, UNITER \citep{chen2020uniter} and VisualBERT \citep{li2020visualbert} are single-stream models with a single Transformer while ViLBERT \citep{lu2019vilbert}, LXMERT \citep{tan2019lxmert}, and ALBEF \citep{ALBEF} are dual-stream models which encode image and textual inputs separately before fusing them. These models all use masked language modelling and image-text matching objectives for pretraining, with LXMERT additionally pretraining on VQA and ALBEF using a contrastive loss to align the image and language representations. UNITER, VisualBERT, and LXMERT use a frozen Faster R-CNN \citep{ren2015faster} to extract region-based features from the image, while ALBEF directly encodes the image with a Vision Transformer \citep{dosovitskiy2020image}.
 
\paragraph{Cross-modal transfer.}
 
Prior work has found models trained on multimodal data to perform better on unimodal downstream tasks than models trained only on one modality. \citet{zadeh2020foundations} found models trained on multimodal input to perform better than text-only models on three NLP tasks, while \citet{testoni2019quantifiers} showed that models trained on textual, visual, and auditory input were better at a quantification task than models trained only on a single modality. Using a task involving queries about typical colours of objects, \citet{norlund2021transferring} found BERT trained on linguistic and visual features to outperform BERT trained on language data filtered for mentions of colour. \citet{frank2021visionandlanguage} investigated the cross-modal alignment of pretrained V+L models with an ablative method based on masked modelling. \citet{lu2021cigli} propose an image--text fusion model to solve a novel image generation task from a textual description and an image prompt based on NLVR2.

\paragraph{Summary.}

The datasets commonly used to evaluate V+L models such as VQA and NLVR2 lack fine-grained interpretability, due to the lack of annotations for semantic phenomena involved in each example. Additionally, multiple semantic phenomena co-occur within a single example, making it difficult to control the training distribution and assess the generalisation abilities of models. Furthermore, existing V+L datasets only present the scene in the visual modality and cannot be used to evaluate a V+L model's cross-modal transfer ability. 

Existing synthetic datasets (e.g. \textsc{ShapeWorld}, CLEVR) fail to split the training and testing distributions along a dimension relevant to the specific task, because they generate captions based on randomly generated images. Unlike existing datasets, \travlr\ fully exploits the benefits of a synthetic dataset by strictly controlling the training and evaluation distributions to test the generalisation abilities of V+L models and avoid statistical biases from language priors and non-uniform distributions. 

\section{\travlr: Cross-Modal Transfer of Visio-Linguistic Reasoning}

We construct \travlr, a synthetic dataset comprising four visio-linguistic reasoning tasks: spatiality, cardinality, quantifiers and numerical comparison. These tasks were previously identified to be challenging for text-only models \citep{lin2021fast, dua2019drop, ravichander2019equate}. \travlr\ aims to evaluate the extent to which pretrained V+L models already encode or are able to learn these four relations between entities present in input scenes. We first describe the general task format, and then describe each task.

Given a scene with objects, $S=\{o_1, ..., o_n\}$, where each object can be represented as a tuple $<$~\textit{colour, shape, position}~$>$, and a textual query $q$ involving some relation $r(o_1, ..., o_i)$ between two or more objects in $S$, each task involves learning a function $y = f(S, q)$ where $y \in \{\textit{true}, \textit{false}\}$. This is essentially a binary classification task. For instance, in the spatiality task, the relation $r$ (e.g., \textit{above}) compares the positions of two objects. In the numerical comparison task, noun phrases in the query refer to subsets of objects, while the relations (e.g., \textit{more}) compare the cardinality of two sets of objects. Assigning a truth value to the query thus involves reasoning over several objects. 

However, a model can never have direct access to the underlying representation scene and must operate on visual or textual forms, and $S$ may be represented in the form of an image or a textual description. In prior work such as VQA, $S$ is presented as an image. In \travlr, $S$ is represented bimodally as an $<$ \textit{image, caption} $>$ pair.

Each example consists of an image, an accompanying caption, and a query. Images include abstract objects arranged in a $6\times6$ grid. We draw from five colours and seven shapes, giving 35 unique objects in total. The shapes used in \travlr\ are from the Visual Genome \citep{krishna2017visual}, a commonly-used pretraining dataset. Each caption fully describes the image with the coordinates of each object (e.g., ``There is a red circle at A 1, a blue square at B 2...''). A description of the coordinate system, e.g., ``Columns, left to right, are ordered A to F. Rows, top to bottom, are ordered 1 to 6.'' is prepended to the caption. The caption and query are separated by the \texttt{[SEP]} token when presented to the models. Removing the caption reduces our tasks to VQA-like tasks.

\vspace{-0.2em}
\subsection{Novel Evaluation Settings}

Encoding the scene as both an image and a caption allows models to be trained and evaluated on a combination of three settings: i) image-only, ii) caption-only, and iii) both image and caption input settings. The query is presented as part of the text input in each setting. In the caption-only setting, a blank white image is presented to the models. 

\vspace{-0.2em}
\subsection{Reasoning Tasks} \label{section:tasks}

In contrast to existing synthetic datasets (e.g. \textsc{ShapeWorld} and CLEVR), we do not generate queries post-hoc based on pre-generated scenes, and instead generate scenes constrained along a task-relevant dimension. For instance, in generating the training and out-of-distribution (OOD) test sets for the spatial relationship task, we ensure that the positions of the queried objects do not overlap between the training and test sets \emph{along the relevant axis} (e.g. the horizontal axis for horizontal relations \textit{left}/\textit{right}). We indicate the train/test splits based on \textit{pairs} in angled brackets.  

\vspace{-0.2em}
\paragraph{Spatiality.} 
This task involves queries of the form ``The \texttt{[object1]} is \texttt{[relationship]} the \texttt{[object2]}.'' (e.g., ``The red circle is to the right of the blue square.''). The possible relationships are \textit{to the left of}, \textit{to the right of}, \textit{above}, \textit{below}. For horizontal relationships (left/right), the train and test sets are split based on the pair $<$\texttt{column(object1),} \texttt{column(object2)}$>$ (see Appendix \ref{sec:datasetfig}), while for vertical relationships (above/below), the train--test split is based on the pair $<$\texttt{row(object1),}\texttt{row(object2)}$>$. This tests the model's ability to generalise its understanding of spatial relationships along the relevant dimension, as opposed to memorising fixed positions. 

\paragraph{Cardinality.}
This task involves queries of the form ``There is/are \texttt{[number]} \texttt{[shape]} object(s).'' (e.g., ``There are 3 circle objects''). The train and test sets are split by the $<$\texttt{number, shape}$>$ pair occurring in the input image/caption; e.g., instances containing 2 circles and 3 triangles could occur in the training distribution, while instances with just 3 circles occur only in the OOD test distribution. 

\vspace{-1mm}
\begin{table}[ht!]
\small
    \centering
    \begin{tabular}{l|l}
         All & $<\texttt{[attr1]}\cap \texttt{[attr2]}, \texttt{[attr2]} \setminus \texttt{[attr1]}>$ \\
        Not all & $<\texttt{[attr1]}\cap \texttt{[attr2]}, \texttt{[attr1]} \setminus \texttt{[attr2]}>$ \\
        No & $<\texttt{[attr1]}\setminus \texttt{[attr2]}, \texttt{[attr2]} \setminus \texttt{[attr1]}>$ \\
        Some & $<\texttt{[attr1]}\setminus \texttt{[attr2]}, \texttt{[attr1]} \cap \texttt{[attr2]}>$ \\
        Only & $<\texttt{[attr1]}\cap \texttt{[attr2]}, \texttt{[attr1]} \setminus \texttt{[attr2]}>$ \\
        Not only & $<\texttt{[attr1]}\cap \texttt{[attr2]}, \texttt{[attr2]} \setminus \texttt{[attr1]}>$
    \end{tabular}
    \caption{Pairs for each quantifier.}
    \label{tab:quantifiers}
    \vspace{-2mm}
\end{table}

\vspace{-1.5em}
\paragraph{Quantifiers.}

This task involves queries of the form ``\texttt{[quantifier]} the \texttt{[attr1]} objects are \texttt{[attr2]} objects.'', where the quantifiers include \textit{all}, \textit{some}, \textit{only} and their negated counterparts \textit{not all}, \textit{none} and \textit{not only}. The train--test split is performed based on the pair $<a, b>$, which varies based on the quantifier, as given in Table \ref{tab:quantifiers}. For instance, for the relationship \textit{not all}, $a$ is the number of objects which fulfil both \texttt{[attr1]} and \texttt{[attr2]}, and $b$ is the number of objects which fulfil \texttt{[attr1]} but not \texttt{[attr2]} (see Appendix \ref{sec:datasetfig}).

\vspace{-0.2em}
\paragraph{Numerical comparison.}

This task involves queries of the form ``There are \texttt{[more/fewer]} \texttt{[attr1]} objects than \texttt{[attr2]} objects.'' (e.g., ``There are more circles than squares.''). The train and test sets are split by the pair $<a, b>$ where $a$ is the number of \texttt{[attr1]} objects, and $b$ is the number of \texttt{[attr2]} objects. Instances for which $|a-b|$ is smaller than a threshold is assigned to the training distribution, and the remaining pairs are assigned to testing. 
$|a-b| \in {1,3}$ and both $a$ and $b$ have a maximal value of 9.
Success in this task is indicative of generalisation based on an understanding of numeral scales and transitivity of comparison; i.e., $a > b$ and $b > c$ implies $a > c$. 

\begin{table*}[ht!]
\setlength{\tabcolsep}{0.15em}
\tablesize
\begin{center}
\begin{tabular}{ @{} p{1em}|c|ccc>{\centering\arraybackslash}p{1.3em}
|ccc>{\centering\arraybackslash}p{1.3em}
|ccc>{\centering\arraybackslash}p{1.3em} @{} } 
  \toprule
  & \textbf{\uline{Train}} 
  & \multicolumn{4}{c|}{\textbf{\uline{Image}}}
  & \multicolumn{4}{c|}{\textbf{\uline{Caption}}}
  & \multicolumn{4}{c}{\textbf{\uline{Image + Caption}}}\\
  
  & \textbf{Test} & \textbf{Image} & \textbf{Caption} & \textbf{Img.\ + Cap.} & \textbf{\#sd} &  \textbf{Image} & \textbf{Caption} & \textbf{Img.\ + Cap.} & \textbf{\#sd} & \textbf{Image} & \textbf{Caption} & \textbf{Img.\ + Cap.} & \textbf{\#sd} \\ 
  \midrule

\multirow{4}{*}{\textbf{S}} &   VisualBERT
& \underline{65.94} (-1.02) & \textbf{48.92} (\textbf{-0.23}) & \underline{\textbf{52.47}} (\textbf{+0.19}) & 3
& \textbf{49.40} (\textbf{+0.09}) & \underline{93.55} (-6.45) & \underline{93.46} (-6.54) & 2
& 49.34 (+0.31) & \underline{70.99} (-1.59) & \underline{71.39} (-1.65) & 3
\\
& UNITER
& \underline{89.67} (+0.96) & 44.36 (+0.48) & 46.15 (-0.19) & 3
& 37.66 (+0.54) & \underline{92.31} (-7.67) & \underline{92.23} (-7.74) & 1
& \textbf{50.15} (\textbf{+0.17}) & \underline{70.75} (+1.24) & \underline{71.17} (+0.40) & 1
\\
& LXMERT
& \underline{\textbf{99.46}} (\textbf{-0.38}) & 38.87 (-0.27) & 48.82 (+0.33) & 3
& 33.52 (+0.47) & 33.52 (+0.47) & 33.52 (+0.47) & 0
& 33.52 (+0.47) & 33.52 (+0.47) & 33.52 (+0.47) & 0
\\
& ALBEF
& 48.28 (+0.10) & 44.54 (-0.26) & 44.85 (-0.18) & 0
& 48.75 (+0.52) & \underline{\textbf{98.42}} (\textbf{-1.58}) & \underline{\textbf{98.42}} (\textbf{-1.58}) & 1
& 48.56 (+0.94) & \underline{\textbf{93.66}} (\textbf{-6.34}) & \underline{\textbf{93.31}} (\textbf{-6.69}) & 2
\\

  \midrule
\multirow{4}{*}{\textbf{C}} &  VisualBERT
& \underline{77.41} (+1.11) & 33.14 (+0.01) & 46.94 (-0.53) & 3
& \textbf{46.63} (\textbf{-0.03}) & \underline{\textbf{99.99}} (\textbf{+0.08}) & \underline{\textbf{99.99}} (\textbf{+0.16}) & 3
& 45.20 (-1.41) & \underline{\textbf{99.94}} (\textbf{+0.10}) & \underline{\textbf{99.94}} (\textbf{+0.18}) & 3
\\
& UNITER
& \underline{77.12} (+0.09) & \textbf{42.36} (\textbf{-0.35}) & 48.55 (+0.23) & 3
& 41.97 (-0.95) & \underline{98.96} (+1.82) & \underline{98.99} (+0.82) & 3
& 42.36 (-1.97) & \underline{98.48} (+0.11) & \underline{98.83} (+1.99) & 3
\\
& LXMERT
& \underline{\textbf{82.90}} (\textbf{-1.66}) & 33.16 (+0.02) & 43.79 (-0.95) & 3
& 45.23 (-0.21) & \underline{60.02} (-18.88) & \underline{60.10} (-18.98) & 3
& \textbf{50.72} (\textbf{+0.34}) & \underline{55.00} (-1.60) & \underline{55.26} (-7.83) & 3
\\
& ALBEF
& \underline{59.19} (+0.43) & 32.67 (-0.05) & \underline{\textbf{53.31}} (\textbf{-0.44}) & 2
& 41.58 (-3.24) & \underline{99.61} (+0.17) & \underline{99.61} (+0.17) & 3
& 43.17 (-2.33) & \underline{99.61} (+0.26) & \underline{99.61} (+0.26) & 3
\\

\midrule

\multirow{4}{*}{\textbf{Q}} & VisualBERT
& \underline{86.59} (-2.73) & 45.09 (+1.16) & \underline{\textbf{60.93}} (\textbf{-2.25}) & 3
& \textbf{49.22} (\textbf{+0.63}) & \underline{\textbf{99.99}} (\textbf{-0.01}) & \underline{\textbf{99.99}} (\textbf{-0.01}) & 3
& \textbf{49.51} (\textbf{-0.16}) & \underline{\textbf{99.98}} (\textbf{-0.01}) & \underline{\textbf{99.98}} (\textbf{-0.02}) & 3
\\
& UNITER
& \underline{95.14} (-1.10) & \textbf{48.89} (\textbf{+0.62}) & \underline{53.99} (-0.13) & 3
& 48.64 (-0.99) & \underline{99.42} (-0.41) & \underline{99.36} (-0.44) & 3
& 48.07 (-1.14) & \underline{97.85} (+2.11) & \underline{98.87} (+1.43) & 3
\\
& LXMERT
& \underline{\textbf{96.72}} (\textbf{-0.94}) & 34.87 (-0.27) & 39.94 (-0.69) & 3
& 43.33 (-0.98) & \underline{92.91} (+6.58) & \underline{90.65} (+7.02) & 3
& 48.95 (-0.14) & 33.93 (-0.01) & \underline{51.02} (+0.17) & 0
\\
& ALBEF
& \underline{66.19} (-0.19) & 43.88 (+0.21) & \underline{54.78} (-1.32) & 3
& 48.27 (+0.19) & \underline{99.98} (+0.12) & \underline{99.98} (+0.12) & 3
& 47.73 (+0.19) & \underline{99.98} (+0.17) & \underline{99.97} (+0.17) & 3
\\

  \midrule
\multirow{4}{*}{\textbf{N}} & VisualBERT
& \underline{58.75} (-14.16) & 40.27 (+1.26) & \underline{\textbf{53.13}} (\textbf{-2.81}) & 3
& \textbf{49.62} (\textbf{-0.42}) & \underline{\textbf{99.77}} (\textbf{-0.06}) & \underline{\textbf{99.75}} (\textbf{+0.00}) & 3
& \underline{\textbf{51.10}} (\textbf{+0.20}) & \underline{\textbf{99.79}} (\textbf{-0.11}) & \underline{\textbf{99.79}} (\textbf{-0.10}) & 3
\\
& UNITER
& \underline{\textbf{63.08}} (\textbf{-22.07}) & 45.52 (+0.81) & \underline{52.97} (-0.62) & 3
& 49.32 (-0.59) & \underline{69.43} (-30.34) & \underline{68.77} (-30.91) & 3
& 46.78 (-1.03) & \underline{64.80} (-34.71) & \underline{63.88} (-35.92) & 3
\\
& LXMERT
& \underline{62.56} (-21.57) & \textbf{45.85} (\textbf{-0.34}) & 46.62 (+0.59) & 3
& 48.62 (-0.94) & \underline{53.10} (-46.30) & \underline{53.15} (-46.21) & 3
& 50.68 (+2.31) & \underline{57.45} (-26.77) & \underline{56.81} (-41.86) & 2
\\
& ALBEF
& 41.93 (-4.30) & 41.23 (-1.27) & 33.88 (-3.13) & 0
& 48.00 (-0.82) & \underline{97.96} (-1.88) & \underline{97.96} (-1.88) & 3
& 47.24 (-1.50) & \underline{98.93} (-0.91) & \underline{98.93} (-0.91) & 3
\\
  \bottomrule
\end{tabular}
\caption{Mean F$_1$ scores on the \travlr\ OOD test sets of four V+L reasoning tasks (\textbf{S:} Spatiality, \textbf{C:} Cardinality, \textbf{Q:} Quantifiers, \textbf{N:} Numerical comparison; relative change from InD results are in parentheses). \#sd indicates the number of converged runs. We report the mean of converged runs if \#sd $>$ 1, and the mean of all runs if \#sd $=$ 0. Above-random results are \uline{underlined}; the result \kj{with the highest mean} in each column is \textbf{bolded}.} 
\label{tab:main_results}
\end{center}
\vspace{-4mm}
\end{table*}

\subsection{Generating \travlr}

To generate examples for each task, we randomly sample object attributes from the distributions defined in \Cref{section:tasks}, ensuring that the \textit{pairs} relevant to each task do not overlap between the train and OOD test sets. We also ensure that all queries in the OOD test set do not occur in the training set. Distractor objects irrelevant to the intended query are finally added to the scene. The spatiality task's training set comprises 32k examples, the training sets of the other tasks comprise 8k examples each due to differences in the amount of data required for convergence in our preliminary experiments.

Prior work on generalisation evaluation recommends the use of in- and out-of-distribution (henceforth InD and OOD, respectively) test sets \citep{csordas2021devil}. Hence, we include validation and InD test sets randomly sampled from the training distribution (10k examples each), in addition to the OOD test set described above (20k examples). We report further dataset statistics in Appendix \ref{sec:dataset}.
% \Cref{tab:sizes} summarises these statistics.

\vspace{-0.3em}
\section{Experiments}
\vspace{-0.2em}
\paragraph{Models.}

We perform experiments on two single-stream models, VisualBERT and UNITER, and two dual-stream models, LXMERT and ALBEF. We use \citet{li2020comparison}'s implementation of VisualBERT, LXMERT, and UNITER, and the original implementation of ALBEF. The image features of LXMERT and UNITER are 36 regions of interest extracted using \citet{tan2019lxmert}'s implementation of a pretrained Faster R-CNN \citep{ren2015faster,Anderson2017up-down}. We use a Detectron model \citep{Detectron2018} to extract image features for VisualBERT. We also train two text-only models, BERT \citep{devlin2019bert} and RoBERTa \citep{liu2019roberta}, as caption-only baselines.

\vspace{-0.2em}
\paragraph{Setup.}

We train models on each task for 80 epochs. Following \citet{csordas2021devil}'s finding that early stopping may lead to underestimation of model performance, we do not do early stopping. We use the recommended hyperparameters for fine-tuning ALBEF on SNLI-VE \citep{xie2019visual} and VisualBERT on NLVR2. Learning rates were adjusted downwards for models/tasks where recommended hyperparameters did not lead to convergence. Each experiment is repeated with three random seeds. \Cref{tab:main_results} reports mean results on seeds leading to above-random performance, and mean results on all seeds when none achieve above-random performance. We present further experimental details in Appendix \ref{sec:exdetails}.  

\vspace{-0.2em}
\subsection{Within-Modality Results}
\vspace{-0.2em}
We first discuss the results of testing the model on the modality it was trained on (\Cref{tab:main_results}). In Appendix \ref{sec:datasetsize}, we examine the effect of dataset size on model performance and discuss the results in detail. In Appendix \ref{sec:training_efficiency}, we discuss differences in training duration between models.

\vspace{-0.2em}
\paragraph{Image-only setting.} Generalising across the four tasks, in the image-only setting, LXMERT is the best performing model across all tasks, while ALBEF is consistently the worst performing model. UNITER outperforms VisualBERT in all tasks except the cardinality task. 

\vspace{-0.2em}
\paragraph{Caption-only and image+caption settings.}

Generally, the performance of all models except LXMERT in the caption-only and image+caption settings is better than performance in the image-only setting. Across tasks and models, performance in the caption-only closely resembles performance in image+caption settings, suggesting a strong textual bias when both modalities are presented. In the caption-only and image+caption settings, LXMERT is consistently the worse performing model, and VisualBERT is the best performing model in all but the spatiality task. On the cardinality and quantifiers tasks, all models except LXMERT achieve close to a perfect F$_1$ score, performing similarly to RoBERTa and BERT. However, they are outperformed by RoBERTa when trained on smaller datasets. 

\vspace{-0.2em}
\paragraph{Spatiality.} While UNITER and LXMERT achieve above-random performance with 8k examples, VisualBERT requires at least 16k examples, and ALBEF fails to learn the task on the full 32k dataset. 32k is a significant number of examples given the task's simplicity. For comparison, the full VQA dataset consists of only 443k training examples. Transformer-based models thus face similar issues to CNN and LSTM models, which \citet{johnson2017clevr} found to have trouble learning spatial relationships and often memorise the absolute object positions. 

A potential explanation for the superior performance of UNITER and LXMERT could be that unlike the other models, spatial coordinates from the bounding boxes are explicitly encoded in the input to the image encoders, which the models can directly exploit. This is unavailable to ALBEF, which takes in the image as input directly instead of relying on a separate object detector. VisualBERT does not make use of these spatial coordinates, which may impair its ability to relate the positions of objects. \citet{bugliarello2021multimodal} and \citet{frank2021visionandlanguage} posited this limitation of VisualBERT to explain its poor performance on tasks such as RefCOCO+ and Masked Region classification, but the impact of this limitation on spatial reasoning has hitherto not been directly investigated. LXMERT converges in fewer epochs compared to all other models. LXMERT only required 4 epochs of training on the 32k dataset to exceed F$_1$=90 on the InD test set, while UNITER required about 60 epochs (see further discussion in Appendix \ref{sec:training_efficiency}). 

In the caption-only and image+caption settings, UNITER and ALBEF are notably more sensitive to random seeds and fail to achieve convergence in several cases. Furthermore, performance is poorer in the image+caption than in the caption-only setting across all models. This runs counter to existing expectations that bimodal representation of the same information should improve performance \citep{zadeh2020foundations, testoni2019quantifiers}. Several models exhibit $>5$ point differences in F$_1$ on the InD and OOD test sets. BERT requires at least 8k examples to achieve an F$_1$ score above 60, corroborating findings by \citet{lin2021fast} that BERT requires a significant number of examples to learn a simple natural language inference task. 

\vspace{-0.3em}
\paragraph{Cardinality.}

On the cardinality task, all models achieve non-random performance when trained with 8k examples. Across all settings, performance on the InD and OOD test set is similar, indicating that models are able to generalise to unseen object--number pairs. However, despite the task's simplicity, in the image-only setting, no model achieves above an F$_1$ of 85 when trained on the full dataset, and when trained on 1k examples, all models perform poorly with F$_1$ scores below 60. This corroborates \citet{parcalabescu2020seeing}'s finding that current V+L models face difficulties with counting objects in images. Models are generally successful in the caption-only setting, corroborating \citet{wallace2019nlp}'s findings that numeracy is encoded in the embeddings of language-only models. 

\vspace{-0.3em}
\paragraph{Quantifiers.}

All models perform well on the quantifiers task in most settings, with exceptions that conform to the overall trends described above (e.g. LXMERT in the image+caption setting). Good performance on the OOD dataset indicates that models are not memorising specific numbers of objects and instead use more general strategies for understanding quantifiers. This parallels psycholinguistic findings that comprehension of non-exact quantifiers does not correlate with counting skills in human children \citep{dolscheid2015counting}.

\vspace{-0.3em}
\paragraph{Numerical comparison.}

Unlike the other tasks, there is a significant difference between the InD and OOD settings for the numerical comparison task across all settings. In the image-only setting, all models except ALBEF, achieve F$_1$s above 70 on the InD test set, while achieving much lower F$_1$s (55--65) on the OOD test set. In the caption-only and image+caption settings, while all models achieve close to F$_1$=100 on the InD test set, UNITER and LXMERT do not generalise well to the OOD test set, showing a substantial drop in performance. Our results suggest that models face difficulties generalising beyond the training distribution to unseen number pairs. 

\subsection{Cross-Modal Transfer}

Despite performing well in the within-modality settings, \textbf{none} of the models succeed at performing zero-shot cross-modal transfer to an unseen modality. Nevertheless, given the success of few-shot learning, we may expect few-shot cross-modal transfer to be plausible. We hence conduct few-shot learning experiments on the best performing models that were fine-tuned on the full dataset of the spatiality and quantifiers tasks. We fine-tuned the models on 200 examples of the unseen modality for 20 epochs before testing (see details in Appendix~\ref{sec:fewshot}). Unfortunately, none of the models achieved above random performance even after this additional fine-tuning. We conclude that they can perform neither zero-shot nor few-shot cross-modal transfer, and that existing V+L representation learning methods have yet to succeed at producing truly multimodal (or modality-agnostic) representations.

\vspace{-0.15em}
\subsection{Adding and Dropping Modalities}
\vspace{-0.2em}
We now discuss the effects of either adding or dropping a modality to the input presented during testing. First, models trained in the image+caption setting perform similarly when tested in the image+caption and caption-only settings. However, when tested in the image-only setting, models perform poorly, exhibiting (close to) random performance in most cases. This indicates that all models tend to overly focus on the caption during training. Second, models trained only on captions perform similarly when tested in the image+caption setting. In contrast, testing a model trained only on images in the image+caption setting results in a significant performance drop. In most cases, F$_1$ is close to random, although all models except LXMERT manage to maintain above-random performance on the quantifier task. This could indicate that the models are easily distracted by the textual modality. Together with the general similarity between results in the caption-only and image+caption settings, these results indicate a bias towards the textual modality. 

\section{Further Analyses}

\vspace{-0.2em}
\subsection{The Impact of Pretraining}
\vspace{-0.2em}
We conduct additional experiments on VisualBERT, UNITER and LXMERT trained using the unified \textsc{Volta} \citep{bugliarello2021multimodal} framework, where V+L models were pretrained using the same dataset in a standardised manner. This allows us to understand the extent to which differences in models' performance can be attributed to the pretraining data or objective, rather than model architecture. 
First, \textsc{Volta} LXMERT loses its advantage in both performance and training efficiency over the other models. We suggest that LXMERT's superior performance is due to some aspect of its pretraining, such as the larger size of or presence of VQA examples in its pretraining dataset, or its use of an additional VQA pretraining objective. Furthermore, \textsc{Volta} VisualBERT's performance is significantly improved over the original VisualBERT, likely because it was pretrained with more data. UNITER also outperforms the other two models (also reported by \citet{bugliarello2021multimodal} on RefCOCO+ and NLVR2), which suggests some advantage of its specific architecture.
All three \textsc{Volta} models achieve similar results on the caption-only setting of the 8k cardinality dataset. As all \textsc{Volta} models were initialised with BERT weights, we suggest that the poor performance of the original LXMERT on textual input is due to its lack of initialisation with BERT weights \citep{tan2019lxmert}. We discuss the results in greater detail  in Appendix \ref{sec:volta}.

\vspace{-0.2em}
\subsection{The Impact of Catastrophic Forgetting}
\vspace{-0.2em}
A potential concern with fine-tuning on one modality and testing on another is that the models may overfit to the fine-tuning modality, resulting in the catastrophic forgetting of cross-modal information. To reduce this possibility, we freeze the pretrained representations and only fine-tune the classification layers. In this setting, all models except LXMERT fail to go beyond random performance. Although LXMERT achieves above-random performance in the image-only setting for two tasks, it is still relatively poor (Spatiality: OOD=52.29, InD=52.67; Cardinality: OOD= 56.39, InD=60.88). This indicates that most models require representation fine-tuning to perform reasoning tasks, and it is unlikely that the inability to transfer cross-modally was due to catastrophic forgetting during fine-tuning. 

\vspace{-0.2em}
\subsection{Modal Dropout}
\vspace{-0.2em}
One may be concerned that the presentation of the same information in the image+caption setting facilitates overfitting to either modality. We investigate whether models' textual bias can be overcome by randomly ``dropping out'' either the image or caption input when training in the image+caption setting. We find modal dropout to not significantly alleviate bias towards the textual modality. We discuss our findings in detail in Appendix \ref{sec:mixed}.

\vspace{-0.2em}
\subsection{\kj{Error analysis}}
\label{sec:erroranalysis}
\vspace{-0.2em}

 We now report findings from our error analysis on selected experiments where models converged but did not achieve perfect or near perfect F$_1$ scores, which can reveal trends that explain poor performance. We describe our analysis on the spatiality and cardinality tasks here, referring readers to Appendix~\ref{sec:erroranalysisapp} for details on quantifiers and numerical comparison. 
 
 Across all tasks, errors are intuitive and explainable, and the performance of models is generally poorer on boundary cases. For example, models are more error-prone when objects are positioned close to each other in the spatiality task, and where only one object either falsifies or confirms the query in the quantifier task. On the numerical comparison task, we observe more errors when the difference between the numerals being compared is small.

\subsubsection{Spatiality}
\vspace{-2mm}
\begin{figure}[h!]
\small
    \centering
    \begin{subfigure}{0.45\textwidth}
    \centering
\includegraphics[width=0.78\linewidth]{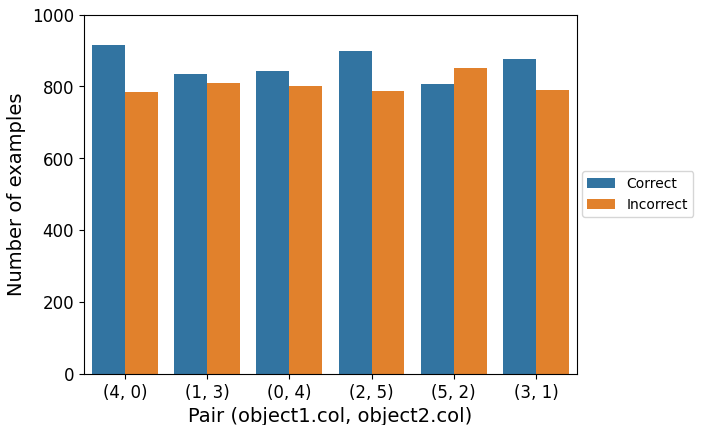}
\caption{Queries involving horizontal relationships, by the pairs of the columns of the queried objects.}
\end{subfigure}
\hspace{\fill}
\begin{subfigure}{0.45\textwidth}
\vspace{0.35em}
\centering
\includegraphics[width=0.78\linewidth]{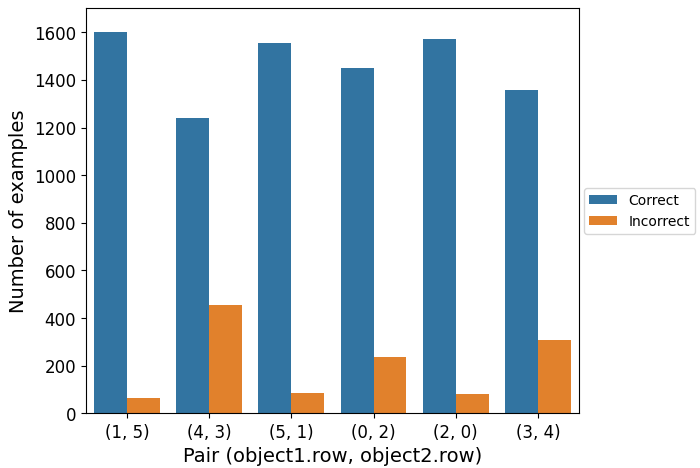}
\caption{Queries involving vertical relationships, by the pairs of the rows of the queried objects.}
\end{subfigure}
    \caption[Error analysis of VisualBERT trained on spatiality task, image-only setting]{Error analysis of VisualBERT trained on spatiality task, image-only setting, tested on OOD test set. (F$_1$=69.78; seed=0)}
    \label{fig:analysis_vbert}
\vspace{-3mm}
\end{figure}

In the spatiality task, VisualBERT was unable to achieve F$_1$ scores above 70 across all three random seeds. Figure~\ref{fig:analysis_vbert} shows that this was due to VisualBERT only correctly answering queries involving vertical relationships, but not horizontal ones. In the figure, the x-axis is organised by the pairs of object coordinates relevant to the query (i.e., columns for horizontal relationship queries, and rows for vertical relationship queries). A similar pattern is observed on all three runs on VisualBERT, and also on UNITER on one random seed. Mediocre results on the spatiality task can thus often be attributed to models successfully learning to answer only a subset of the queries.

\begin{figure}[h!]
\vspace{-1.5mm}
\small
\centering
\centering
\includegraphics[width=0.40\textwidth]{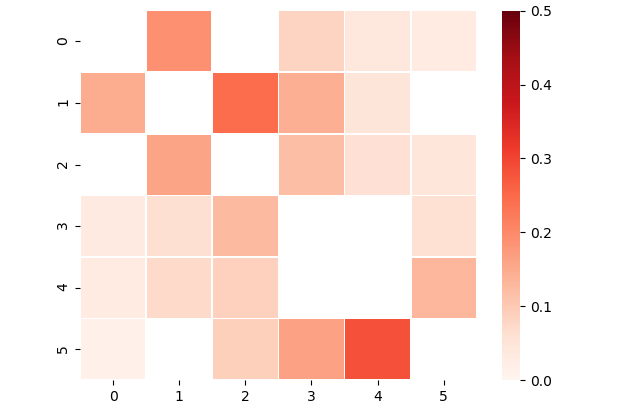}
\vspace{-1mm}
\caption{Percentage of incorrect answers of UNITER in the image-only setting, on InD $<x, y>$ pairs for vertical relationships. (F$_1$=71.79; seed=2)}
\label{fig:analysis-spatial-cm}
\vspace{-2.5mm}
\end{figure}

Figure~\ref{fig:analysis-spatial-cm} shows the percentage of incorrect answers on specific pairs in the InD test set. Note that white entries are pairs which belong to the OOD test set. Overall, models make more mistakes on examples with queried objects positioned closer together (represented by cells closer to the diagonal).

\subsubsection{Cardinality}

\begin{figure}[h!]
\vspace{-2mm}
\small
    \centering
    \includegraphics[width=0.7\linewidth]{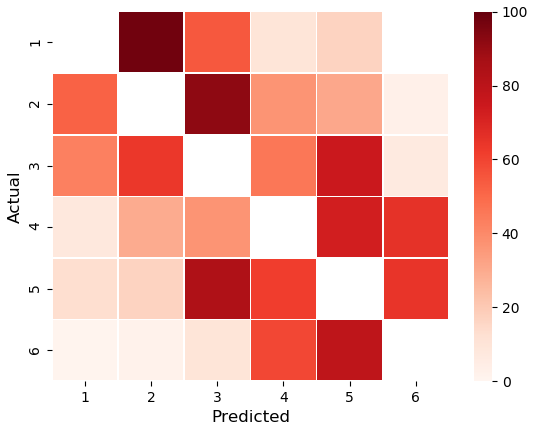}
    \vspace{-2mm}
    \caption[Actual vs. predicted values on the cardinality task.]{Actual vs predicted values (given that predicted answer is true). VisualBERT trained on the cardinality task, image-only setting, tested on InD test set. (F$_1$=77.84; seed=0)}
    \label{fig:analysis_card_2}
    \vspace{-2mm}
\end{figure}

While most models apart from LXMERT succeeded on the cardinality task in the caption-only and image+caption settings, moderate performance was observed in the image-only setting. We found that models performed poorer on images with more objects in the image, inclusive of distractors (refer to Appendix~\ref{sec:erroranalysisapp} for details). Furthermore, models often predict values close to the actual value. Figure~\ref{fig:analysis_card_2} plots the actual number of objects in the scene against the number of objects predicted. Since the task was a true/false task, this analysis is based only on examples for which the model predicted that the query was true. The counting abilities of models thus intuitively resemble that of humans, given that they are confused by a large number of objects in the scene, and are also likely to predict answers close to the actual value when miscounting.

\vspace{-0.3em}

\section{Discussion} \label{section:discussion}
\paragraph{Comparing modalities.} 

As discussed in \S 4.3, V+L models exhibit a clear bias towards the textual modality across single and dual stream models, corroborating findings by \citet{cao2020behind}. This finding also applies to LXMERT, even though it generally performs more poorly on caption than image inputs. Furthermore, the V+L models perform more poorly than unimodal RoBERTa on various tasks in the caption-only setting, similar to \citet{iki2021effect}, who show that V+L pretraining on degrades performance on NLU tasks.

\paragraph{Comparing tasks.} The spatiality task is the hardest task, requiring at least 32k examples to converge in some cases, as opposed to the 8k examples for the other tasks (see Appendix~\ref{sec:datasetsize} for details). Furthermore, convergence of the models on the spatiality task is highly sensitive to random seeds. Among the other tasks, the easiest task is the quantifiers task, followed by cardinality, and finally numerical comparison. In the caption-only and image+caption settings, all models apart from LXMERT achieve a close to perfect F$_1$ on the cardinality and quantifiers tasks. However, on the numerical comparison task, UNITER and LXMERT exhibited a limited ability to generalise outside the training distribution in both image and textual modalities. Thus, while success on the cardinality task indicates that models possess an understanding of the meaning of numbers in absolute terms, the numerical comparison task was able to differentiate models in terms of their understanding of numbers' relative positions on a numeral scale. 

\paragraph{Comparing models.} 

As argued in \S 5.1, the most significant factor differentiating models seems to be their \textbf{pretraining} rather than model architecture. Our findings corroborate \citet{bugliarello2021multimodal}'s findings that differences between models cannot be primarily attributed to differences in model architecture. Unlike LXMERT and ALBEF, VisualBERT and UNITER both succeed on all tasks in all settings. While UNITER generally outperforms VisualBERT in the image-only setting, VisualBERT at times outperforms UNITER in the caption and image+caption settings. The superior performance of VisualBERT in caption-only settings could be due to its comparatively minimal pretraining on V+L objectives, enabling it to retain performance closer to that of text-only models.  

The \textbf{encoding of image features} also has a significant impact on performance in the image-only setting. VisualBERT's poor performance on the spatiality task is likely due to the fact that it does not explicitly encode the spatial coordinates of the bounding boxes in its input. Additionally, it is likely that the pretrained object detector used by every model except ALBEF helped them outperform ALBEF in the image-only setting, despite underperforming it in common V+L benchmarks \citep{ALBEF}. A possible explanation is that the information captured by the Faster R-CNN detector was more relevant to our visio-linguistic tasks. 

Finally, the poor performance of LXMERT on caption-only and image+caption settings reflects its lack of \textbf{initialisation with BERT parameters} before pretraining. Given the superior performance of RoBERTa over BERT, it could be beneficial to initialise models with RoBERTa weights instead.

\section{Conclusion}

\travlr\ allows us to evaluate specific visio-linguistic reasoning skills in isolation, enabling finer-grained diagnosis of model deficiencies. We found some models to learn better from one modality than the other, and some task-setting combinations to be more challenging across the board. Furthermore, existing models may require unreasonably large amounts of data and training steps to learn simple tasks. Improving the sample efficiency and training time of V+L models is a potential direction for future research. Furthermore, while pretrained multilingual models have been shown to demonstrate zero-shot cross-lingual transfer abilities, it is unclear whether V+L models can similarly perform \emph{cross-modal transfer} of downstream task abilities to a modality unseen during fine-tuning. 

We hence contribute \travlr, which enables the evaluation of cross-modal transfer ability by encoding scenes bimodally. We found all models to suffer from a strong textual bias, and an inability to perform zero- and few-shot cross-modal transfer. Given the success of multilingual models with cross-lingual transfer, future work might attempt to bridge the gulf between the visual and textual modalities, perhaps by forcing models to transform images into intermediate representations which more closely resemble language. Visio-linguistic representations capable of such transfer will unlock a whole host of new applications, and we pose this as the next challenge for multimodal modelling.

\section*{Limitations}

We acknowledge that the use of synthetic data is potentially unrealistic given that most applications would require reasoning on real-world input. Nevertheless, we note that because synthetic data is easier, it is still useful as a benchmark for the minimum expected performance of models. We follow the line of argument in prior work (e.g. \citet{johnson2017clevr}) that synthetic datasets not only mitigate the problem of distributional bias in real-world datasets, but also simplify the problem of object recognition to focus on diagnosing genuine abstract reasoning ability. We generally expect V+L models to acquire abstract visual reasoning skills, like humans, rather than being limited to the realm of photographs. \travlr's focus on simple shapes also makes it a test of this ability.

Our experiments on \textsc{Volta} should be taken as preliminary, as a full replication of all experiments was not performed due to resource limitations. Future work could investigate the extent to which the superior performance of \textsc{Volta} UNITER is robust across all tasks, which would indicate a genuine advantage due to UNITER's architecture. 
Furthermore, although we have attributed the superior performance of LXMERT to some aspect of its pretraining, we are unable to pinpoint whether the advantage is specifically due to the large size of its pretraining dataset, its use of VQA examples in the pretraining, or its use of a VQA pretraining objective, and leave this to future work.

Finally, while all models were shown to succeed at generalising to the OOD test setting with varying degrees of success, only the numerical comparison task poses a significant challenge in terms of OOD generalisation. This is expected, since we did not intend to challenge the models with generalisation settings which they are unlikely to encounter in test (i.e., more realistic datasets designed for fine-tuning models to perform real-world applications). \kj{Furthermore, the simplicity of our reasoning tasks enables us to evaluate models' abilities on specific semantic phenomena in isolation.} Nevertheless, future work can extend \travlr\ to include more challenging generalisation tasks, such as generalisation of the meaning of negation over unseen quantifiers, compositional queries which combine multiple semantic phenomena \kj{(e.g. combining propositions via conjunction or disjunction)}, and queries which contain instructions to manipulate the scene.

\section*{Ethics Statement}
\travlr\ is a synthetically generated dataset, and hence poses no ethical issues relating to the source of the data. Its use of abstract shapes has the further advantage of avoiding biases relating to human subjects. 

\bibliography{custom}
\bibliographystyle{acl_natbib}

\clearpage
\appendix

\section{Dataset Details}
\label{sec:dataset}
We now describe the version of \travlr\ used in our experiments, although our dataset generation scripts allow parameters such as the object properties and grid size to be modified. \travlr\ draws from five colours (red, blue, green, yellow, orange) and seven shapes (square, circle, triangle, star, hexagon, octagon, pentagon), yielding a total of thirty-five unique objects. The seven shapes and colours used in \travlr\ are present in the Visual Genome dataset \citep{krishna2017visual}. We use a six by six grid, which allows a maximum of 36 objects in a scene. Each example consists of an image, a caption and a query.

\subsection{Dataset Figures} \label{sec:datasetfig}

\vspace{-5mm}
\begin{figure}[h]
\small
    \centering
    \includegraphics[width=\linewidth]{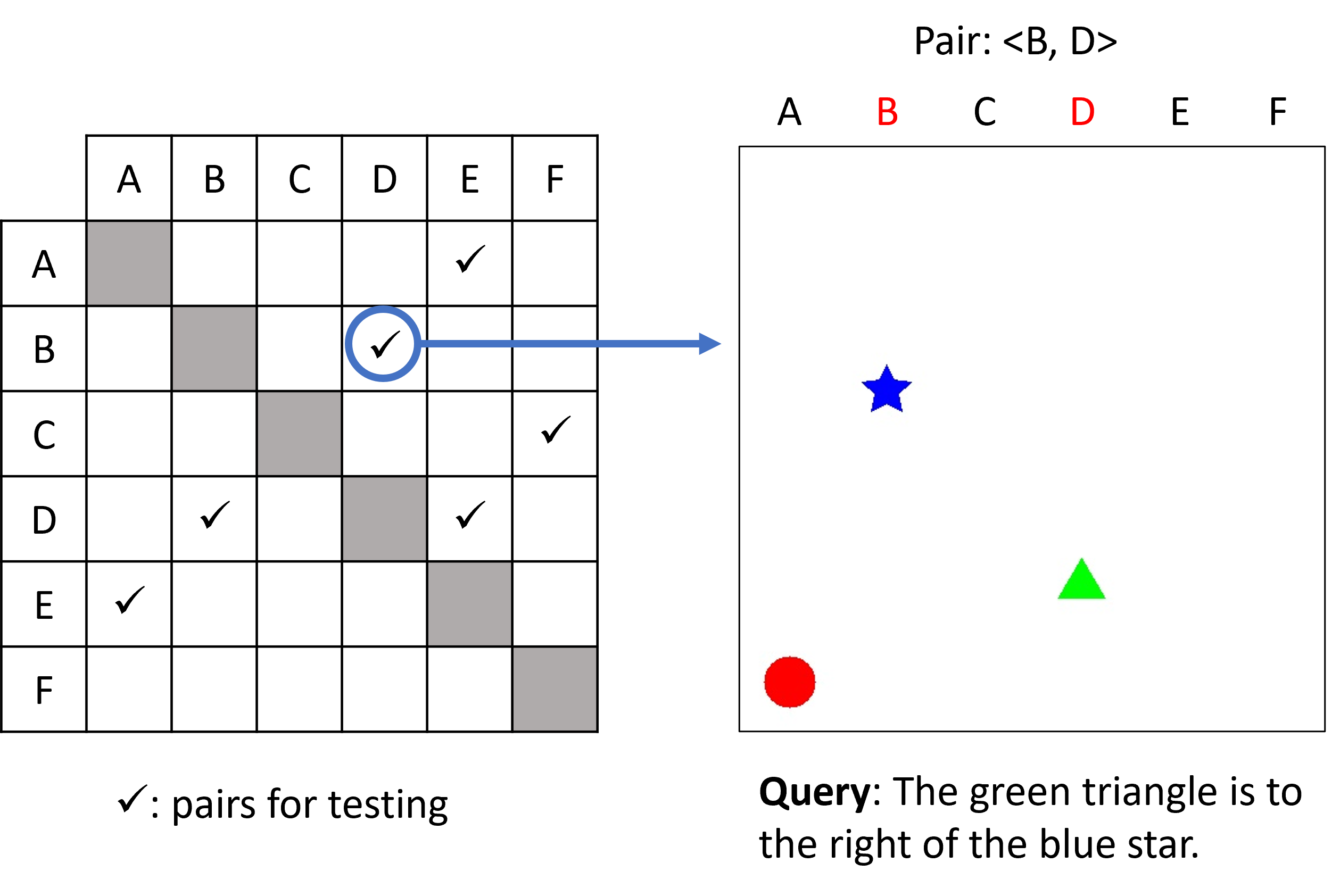}
    \vspace{-3mm}
    \caption{An example of OOD test set construction. In a left/right relationship reasoning task, the relevant dimension is the column ID. Specific ID pairs (\checkmark) are held out to form this test distribution.}
    \label{fig:split}
    \vspace{-5mm}
\end{figure}

\begin{figure}[h]
    \centering
    \includegraphics[width=\linewidth]{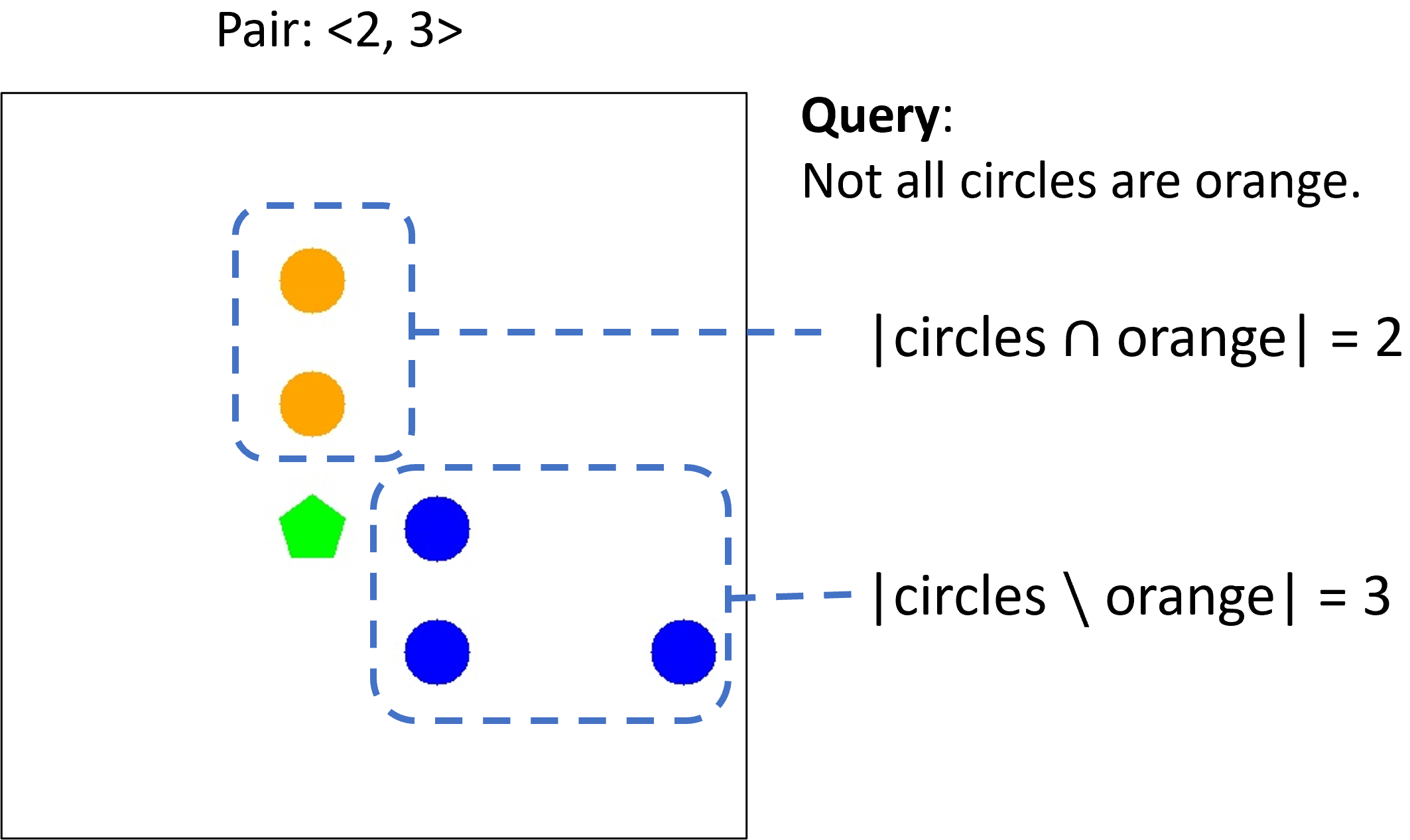}
    \caption{Example instance for \textit{not all} quantifier with pair $<2, 3>$.}
    \label{fig:notall}
    \vspace{-5mm}
\end{figure}

\subsection{Dataset Statistics}

Table \ref{tab:sizes} shows the label distributions (true/false) for the various datasets: training, validation, In-Distribution (InD) test, and Out-Of-Distribution (OOD) test sets. The spatiality dataset is the largest dataset comprising 32k examples, while the other datasets comprise 8k examples. In the remainder of the section, we detail crucial statistics for the splits between the training and OOD test sets.

\begin{table}[!h]
\centering
\tablesize
\setlength{\tabcolsep}{0.6em}
  \begin{tabular}{l|cccc} 
  \toprule
  \textbf{Task} & \textbf{Train} & \textbf{Val.} & \textbf{InD Test} & \textbf{OOD Test}\\
  \midrule
  Spatial & 15903 / 16097 & 4979 / 5021 & 5064 / 4936 & 9918 / 10082 \\
  Cardinality & 4040 / 3960 & 4927 / 5073 & 5043 / 4957 & 10079 / 9921 \\
  Quantifier & 4006 / 3994 & 5003 / 4997 & 5030 / 4970 & 10029 / 9971 \\
  Comparison & 4088 / 3912 & 4926 / 5074 & 4992 / 5008 & 10033 / 9967 \\
  \bottomrule
  \end{tabular}
  \caption{Dataset statistics (no. of True / False).} \label{tab:sizes}
  \vspace*{-2mm}
\end{table}

\vspace{-2mm}

\begin{figure}[h!]
\small
\centering
\begin{subfigure}
{0.5\textwidth}
\begin{subfigure}{0.48\textwidth}
\includegraphics[width=\textwidth]{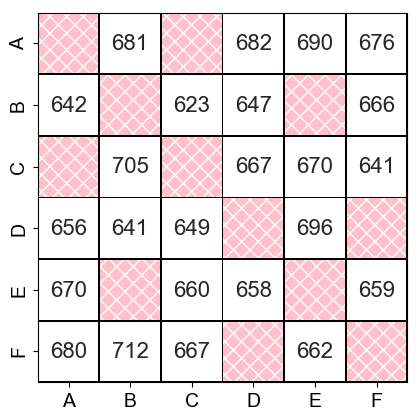}
\caption*{Horizontal Relationship}
\end{subfigure}
% \hspace{\fill}
\begin{subfigure}{0.48\textwidth}
% \vspace{0.35em}
\includegraphics[width=\textwidth]{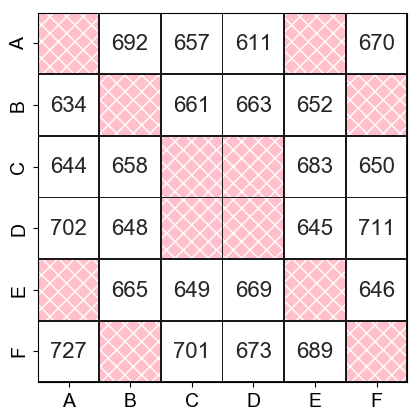}
\caption*{Vertical Relationship}
\end{subfigure}
\caption{Training Set.}
\end{subfigure}

\begin{subfigure}
{0.5\textwidth}
\vspace{0.7em}
\begin{subfigure}{0.48\textwidth}
\includegraphics[width=\textwidth]{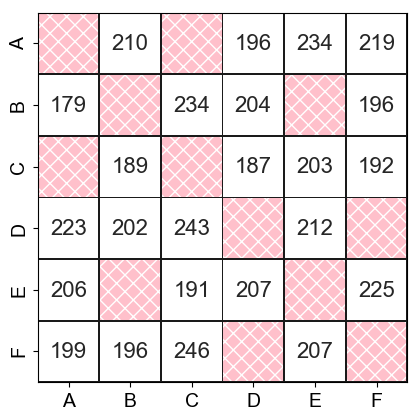}
\caption*{Horizontal Relationship}
\end{subfigure}
% \hspace{\fill}
\begin{subfigure}{0.48\textwidth}
% \vspace{0.35em}
\includegraphics[width=\textwidth]{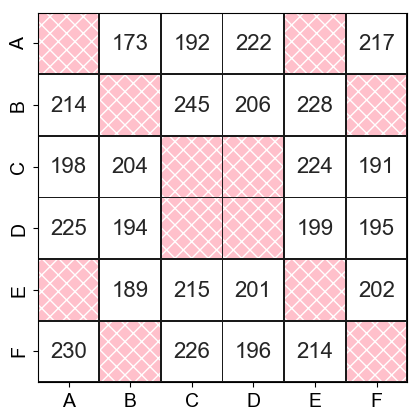}
\caption*{Vertical Relationship}
\end{subfigure}
\caption{In-Distribution Test Set.}
\end{subfigure}

\begin{subfigure}{0.5\textwidth}
\vspace{0.7em}
\begin{subfigure}{0.48\textwidth}
\includegraphics[width=\textwidth]{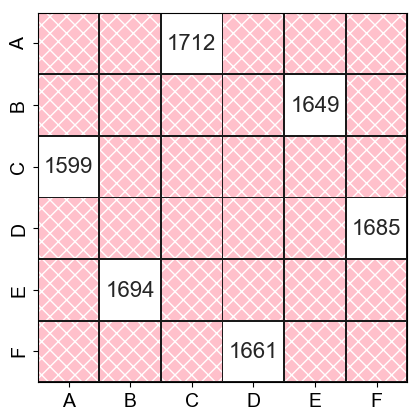}
\caption*{Horizontal Relationship}
\end{subfigure}
% \hspace{\fill}
\begin{subfigure}{0.48\textwidth}
% \vspace{0.35em}
\includegraphics[width=\textwidth]{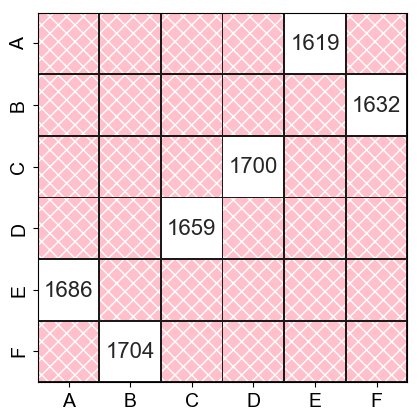}
\caption*{Vertical Relationship}
\end{subfigure}
\caption{Out-of-Distribution Test Set.}
\end{subfigure}
\caption{Spatiality ($<$\texttt{column(object1),} \texttt{column(object2)}$>$ for horizontal relationship, $<$\texttt{row(object1),} \texttt{row(object2)}$>$ for vertical relationship).}
\label{fig:dataset-spatial}
\end{figure}

For the spatiality task, the scene comprises only three objects: the two objects mentioned in the query, and one distractor object. We limit the number of distractors for this task, since we intend the task to evaluate models' ability to compare the position of the objects, rather than their ability to perform object recognition. 

\begin{figure}[t!]
\small
\centering
\begin{subfigure}{0.3\textwidth}
\includegraphics[width=\textwidth]{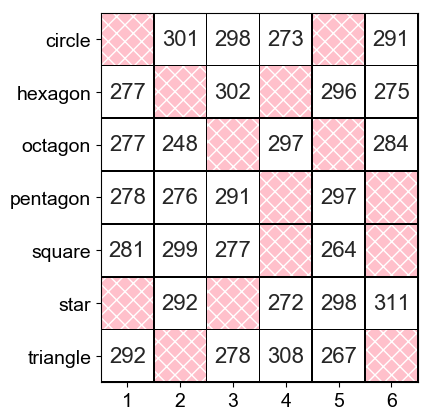}
\caption{Training Set.}
\end{subfigure}

\begin{subfigure}{0.3\textwidth}
\vspace{0.7em}
\includegraphics[width=\textwidth]{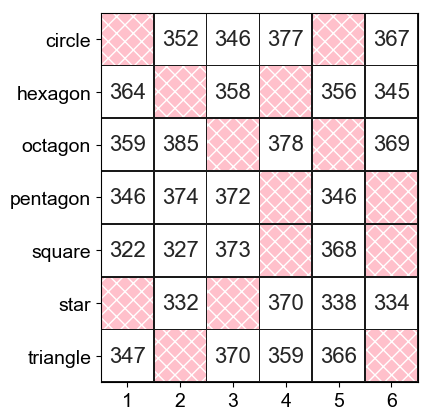}
\caption{In-Distribution Test Set.}
\end{subfigure}

\begin{subfigure}{0.3\textwidth}
\vspace{0.7em}
\includegraphics[width=\textwidth]{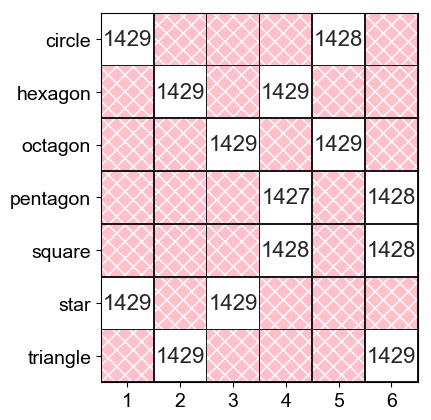}
\caption{Out-of-Distribution Test Set.}
\end{subfigure}
\caption{Cardinality ($<$\texttt{shape, number}$>$).}
\label{fig:dataset-card}
\end{figure}

Figure~\ref{fig:dataset-spatial} shows dataset statistics for the training, InD and OOD spatiality dataset. Each figure indicates the number of examples whose scene instantiates a specific \textit{pair} (this is independent of the query and label). For instance, there are 681 examples instantiating the pair $<A, B>$ for the horizontal relationship. In these examples, the first object in the query is found in Column A, and the second object is found in Column B. As illustrated in Figure~\ref{fig:split}, we ensure no overlap between pairs in the training set and the OOD test set. We note that each example either instantiates a horizontal or vertical relationship, but not both.

Unlike the spatiality task, more objects are present in the remaining tasks, since it is the ability to reason about the number of objects which is under test. For the cardinality task (Figure~\ref{fig:dataset-card}), we limit the number of objects relevant to the query within the range [1, 6], and the maximum number of distractor objects within the range [1, 10]. The relevant pairs shown in Figure~\ref{fig:dataset-card} are $<$\texttt{shape, number}$>$, i.e. the number of objects of that shape.

\begin{figure}[h!]
\small
\centering
\begin{subfigure}{0.3\textwidth}
\includegraphics[width=\textwidth]{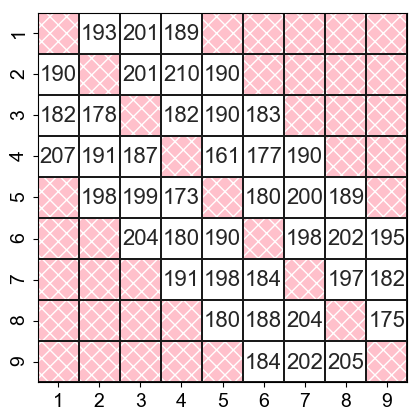}
\caption{Training Set.}
\end{subfigure}

\begin{subfigure}{0.3\textwidth}
\includegraphics[width=\textwidth]{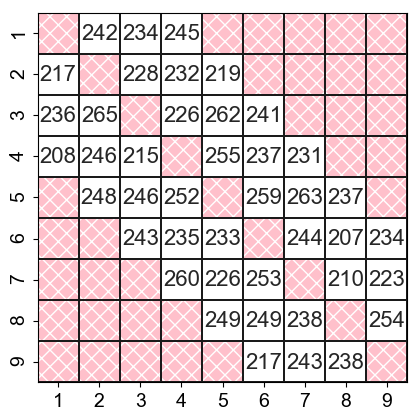}
\caption{In-Distribution Test Set.}
\end{subfigure}

\begin{subfigure}{0.3\textwidth}
\includegraphics[width=\textwidth]{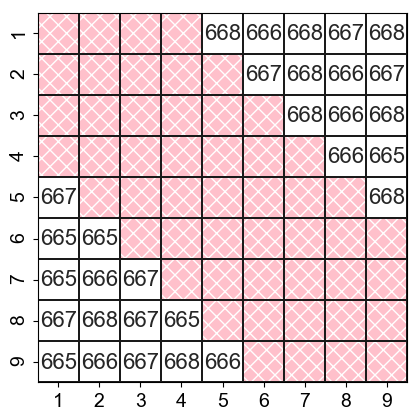}
\caption{Out-of-Distribution Test Set.}
\end{subfigure}
\caption{Numerical Comparison.}
\label{fig:dataset-compare}
\end{figure}

For the numerical comparison task (Figure~\ref{fig:dataset-compare}), we split the train and test sets by the pair $<a, b>$ where $a$ is the number of \texttt{[attr1]} objects, and $b$ is the number of \texttt{[attr2]} objects. The number of \texttt{[attr1]} and \texttt{[attr2]} objects are limited within the range $[1, 9]$, and the number of distractor objects is in the range $[1, 10]$.

\begin{table}[h!]
\small
    \centering
    \begin{tabular}{l|l}
         All & $<\texttt{[attr1]}\cap \texttt{[attr2]}, \texttt{[attr2]} \setminus \texttt{[attr1]}>$ \\
        Not all & $<\texttt{[attr1]}\cap \texttt{[attr2]}, \texttt{[attr1]} \setminus \texttt{[attr2]}>$ \\
        No & $<\texttt{[attr1]}\setminus \texttt{[attr2]}, \texttt{[attr2]} \setminus \texttt{[attr1]}>$ \\
        Some & $<\texttt{[attr1]}\setminus \texttt{[attr2]}, \texttt{[attr1]} \cap \texttt{[attr2]}>$ \\
        Only & $<\texttt{[attr1]}\cap \texttt{[attr2]}, \texttt{[attr1]} \setminus \texttt{[attr2]}>$ \\
        Not only & $<\texttt{[attr1]}\cap \texttt{[attr2]}, \texttt{[attr2]} \setminus \texttt{[attr1]}>$
    \end{tabular}
    \caption{Pairs for each quantifier.}
    \label{tab:quantifiers-app}
    \vspace{-2mm}
\end{table}

\begin{figure}[ht!]
\small
\centering
\begin{subfigure}{0.50\textwidth}
\centering
\begin{subfigure}{0.35\textwidth}
\includegraphics[width=\textwidth]{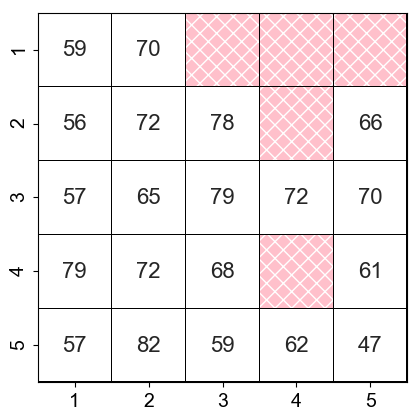}
\caption*{All}
\end{subfigure}
% \vspace{0.15em}
\begin{subfigure}{0.35\textwidth}
\includegraphics[width=\textwidth]{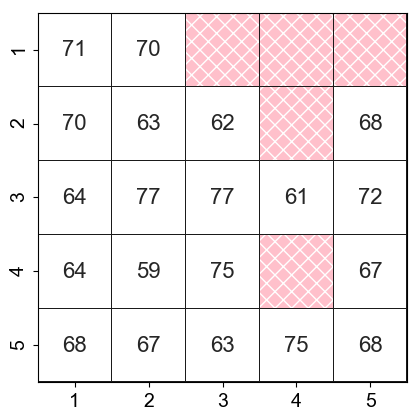}
\caption*{Not All}
\end{subfigure}
\begin{subfigure}{0.35\textwidth}
\vspace{0.35em}
\includegraphics[width=\textwidth]{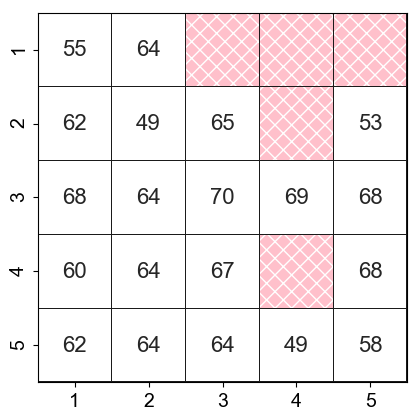}
\caption*{Some}
\end{subfigure}
\begin{subfigure}{0.35\textwidth}
\vspace{0.35em}
\includegraphics[width=\textwidth]{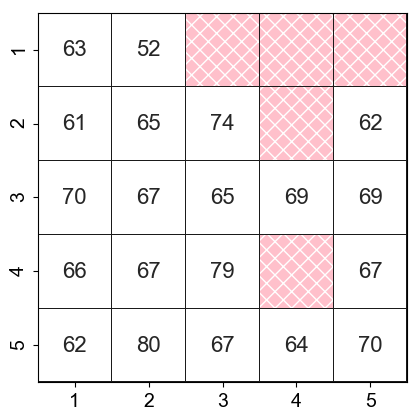}
\caption*{None}
\end{subfigure}
% \hspace{\fill}
\begin{subfigure}{0.35\textwidth}
\vspace{0.35em}
\includegraphics[width=\textwidth]{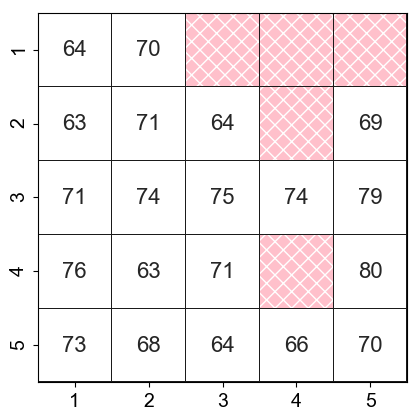}
\caption*{Only}
\end{subfigure}
\begin{subfigure}{0.35\textwidth}
% \vspace{0.35em}
\includegraphics[width=\textwidth]{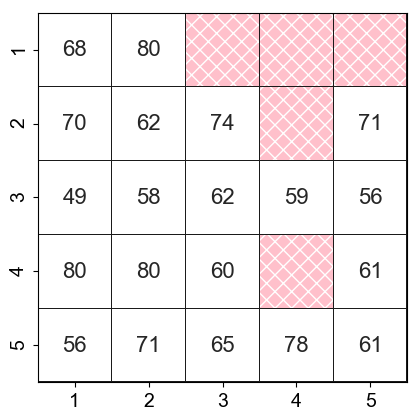}
\caption*{Not Only}
\end{subfigure}
\end{subfigure}
\caption{Quantifiers Training Set.}
\label{fig:dataset-quant}
\end{figure}

For the quantifiers task (Figure \ref{fig:dataset-quant}), we split the training and OOD distributions based on the pairs as stated in Table~\ref{tab:quantifiers-app}, which differ depending on the specific quantifier. We limit the number of objects relevant to each part of the pair within the range $[1, 5]$, and the number of distractor objects within the range $[2, 8]$. 

\begin{figure}[ht!]
\small
\centering
\begin{subfigure}
{0.50\textwidth}
\centering
\begin{subfigure}{0.35\textwidth}
\includegraphics[width=\textwidth]{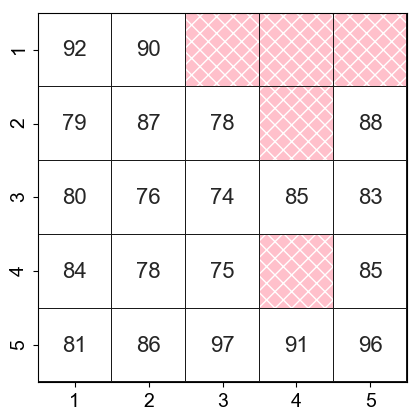}
\caption*{All}
\end{subfigure}
% \vspace{0.15em}
\begin{subfigure}{0.35\textwidth}
\includegraphics[width=\textwidth]{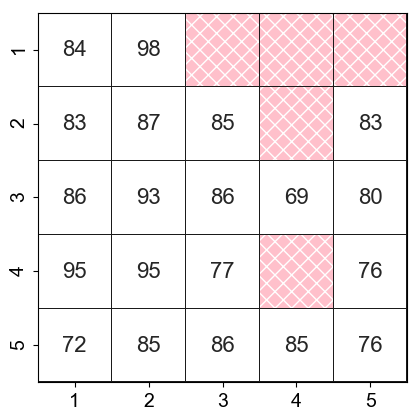}
\caption*{Not All}
\end{subfigure}
\begin{subfigure}{0.35\textwidth}
\vspace{0.35em}
\includegraphics[width=\textwidth]{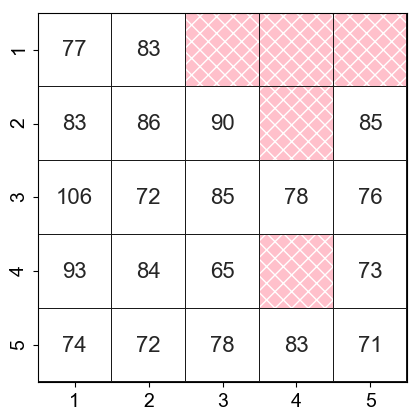}
\caption*{Some}
\end{subfigure}
\begin{subfigure}{0.35\textwidth}
\vspace{0.35em}
\includegraphics[width=\textwidth]{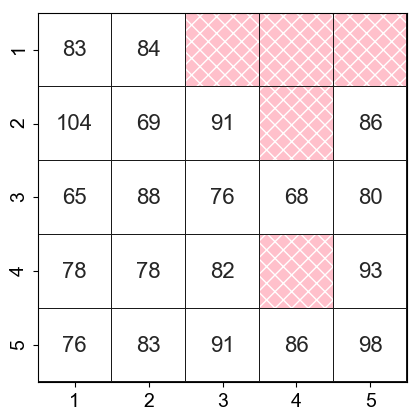}
\caption*{None}
\end{subfigure}
% \hspace{\fill}
\begin{subfigure}{0.35\textwidth}
\vspace{0.35em}
\includegraphics[width=\textwidth]{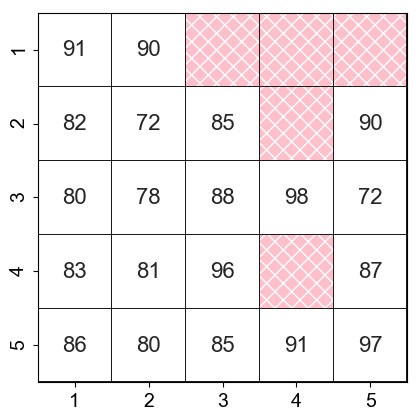}
\caption*{Only}
\end{subfigure}
\begin{subfigure}{0.35\textwidth}
% \vspace{0.35em}
\includegraphics[width=\textwidth]{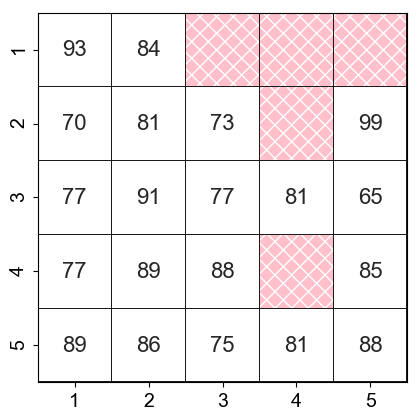}
\caption*{Not Only}
\end{subfigure}
\caption{In-Distribution Test Set.}
\end{subfigure}

\begin{subfigure}
{0.50\textwidth}
\centering
\begin{subfigure}{0.35\textwidth}
\includegraphics[width=\textwidth]{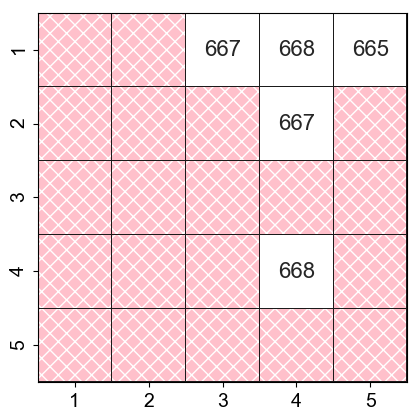}
\caption*{All}
\end{subfigure}
% \vspace{0.15em}
\begin{subfigure}{0.35\textwidth}
\includegraphics[width=\textwidth]{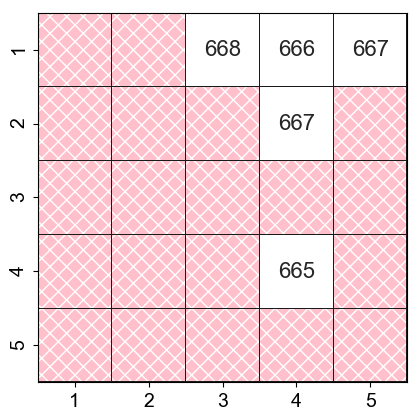}
\caption*{Not All}
\end{subfigure}
\begin{subfigure}{0.35\textwidth}
\vspace{0.35em}
\includegraphics[width=\textwidth]{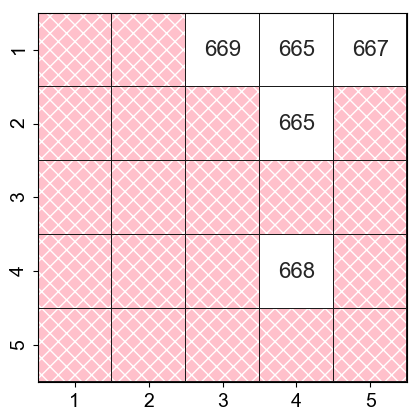}
\caption*{Some}
\end{subfigure}
\begin{subfigure}{0.35\textwidth}
\vspace{0.35em}
\includegraphics[width=\textwidth]{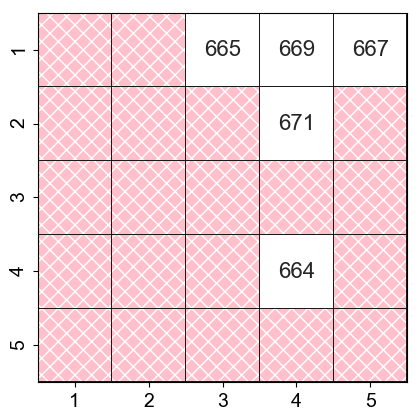}
\caption*{None}
\end{subfigure}
% \hspace{\fill}
\begin{subfigure}{0.35\textwidth}
\vspace{0.35em}
\includegraphics[width=\textwidth]{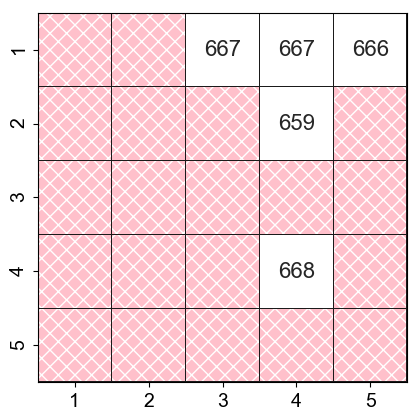}
\caption*{Only}
\end{subfigure}
\begin{subfigure}{0.35\textwidth}
% \vspace{0.35em}
\includegraphics[width=\textwidth]{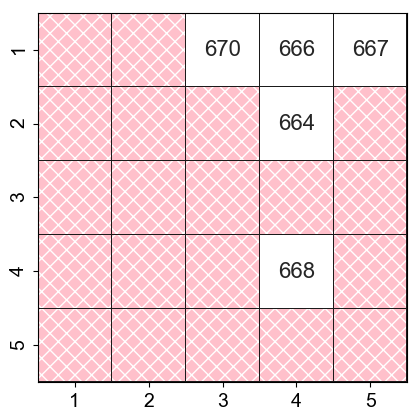}
\caption*{Not Only}
\end{subfigure}
\caption{In-Distribution Test Set.}
\end{subfigure}

\caption{Quantifiers Test Sets.}
\label{fig:dataset-quant}
\end{figure}

\FloatBarrier
\section{Experiment details}
\label{sec:exdetails}
We conducted the following fine-tuning experiments:

\begin{itemize}[leftmargin=1em]
\itemsep0em 
    \item 4 V+L models on all 4 tasks on 3 modality settings, on 3 random seeds each = 144 main experiments.
    \item 2 text-only models on all 4 tasks, on 3 random seeds each = 24 experiments.
    \item Replications of the above experiments on 3 smaller subsets of the original dataset, on a single random seed = 144 + 24 = 168 experiments (see Appendix \ref{sec:datasetsize}).
    \item All models on all 4 tasks with frozen pretrained representations.
    \item 4 V+L models on 4 tasks on the modal dropout/``mixed'' setting (see Appendix \ref{sec:mixed}).
    \item Selected experiments on the \textsc{Volta} implementation of 3 V+L models (see Appendix \ref{sec:volta}).
    \item Selected experiments using the few-shot learning setting (see Appendix \ref{sec:fewshot}).
\end{itemize}

\subsection{Metric}

The metric used is the macro F$_1$ score, which is computed as the arithmetic mean of the F$_1$ score of each of the two classes (true and false). We use F$_1$ instead of accuracy due to small imbalances in the classes (see Table~\ref{tab:sizes}).

\begin{equation}
    F_1 = 2 * \frac{Precision * Recall}{Precision +Recall}
\end{equation}

\subsection{Hyperparameters}

We train models for 80 epochs without early stopping. As the hyperparameters recommended for fine-tuning UNITER and LXMERT did not lead to convergence on some tasks, we adjusted learning rates downwards until the model converged. For LXMERT, we used a learning rate of 5e-5 only for the spatiality task, and 5e-6 elsewhere. 

\begin{table}[!h]
\centering
\small
  \begin{tabular}{l|cccc} 
  \toprule
  \textbf{Model} & \textbf{Batch Size} & \textbf{Learning rate} \\
  \midrule
  VisualBERT & 64 & 2e-5 \\
  UNITER & 32 & 5e-6 \\
  LXMERT & 32 & 5e-6 / 5e-5\\
  ALBEF & 256 & 2e-5\\
  VOLTA VisualBERT & 32 & 5e-6 \\
  VOLTA UNITER & 32 & 5e-6\\
  VOLTA LXMERT & 32 & 5e-6\\
  \bottomrule
  \end{tabular}
  \caption{Fine-tuning hyperparameters.} \label{tab:hyperparams}
\end{table}

\subsection{Runtimes}
Table \ref{tab:runtimes} provides an estimate of the runtimes for fine-tuning each model on different dataset sizes in different settings. Experiments were run either on NVIDIA GeForce RTX 2080 Ti or NVIDIA TITAN RTX GPUs. The longest experiments on ALBEF with the 32k dataset are estimated to lead to carbon emissions of 15.24 kgCO$_2$eq \citep{lacoste2019quantifying}.

\begin{table}[h!]
\setlength{\tabcolsep}{0.4em}
\small
\begin{center}
\begin{tabular}{ @{} >{\centering\arraybackslash}p{13mm}|c|c|c|c @{} } 
    \midrule
     \textbf{Dataset size} & \textbf{Model}
   & \textbf{Image} & \textbf{Caption}
   & \textbf{Img.+Cap.} \\
  \midrule
  \multirow{6}{*}{32k} & UNITER & 15 & 27 & 27 \\
   &   VisualBERT & 15 & 27 & 27 \\
   &     LXMERT & 25.5 & 36.5 & 36.5\\
   & ALBEF & 38 & 38 & 38\\
   & BERT & - & 12 & - \\
    &  RoBERTA	 & - & 12 & -\\
    \midrule
    
  \multirow{6}{*}{8k}  & UNITER	& 3.5 & 7 & 7 \\ 
& VisualBERT & 5 & 7 & 7 \\
& LXMERT	& 6.5 & 10.5 & 10.5 \\
 & ALBEF & 8 & 9 & 9 \\
& BERT	 & - & 3 & - \\
 & RoBERTA & - & 3 & - \\
 
 \bottomrule
\end{tabular}
\caption{Approximate experiment run times in hours.} 
\label{tab:runtimes}
\end{center}
\vspace{-4mm}
\end{table}

\vspace{-2mm}
\subsection{Model Details}

Table \ref{tab:model_data} summarises the amount and sources of pretraining data used in the pretraining of various models we evaluate. Table \ref{tab:model_params} reports the total trainable parameters for each model.

\begin{table}[!htbp]
\small
\setlength{\tabcolsep}{0.15em}
\begin{center}
\begin{tabular}{ @{} l|>{\centering\arraybackslash}p{20mm}|c|c|c|c|c @{} } 
    \toprule
     \textbf{Model} &  \textbf{Pretraining Dataset Size} & COCO & VG & CC & SBU & Other \\
  \midrule
  VisualBERT & 600k & \checkmark & & & \checkmark & \checkmark \\ 
  UNITER & 5.6M & \checkmark & \checkmark & \checkmark & \checkmark & \\
  LXMERT & 9.18M & \checkmark & \checkmark & \checkmark & \checkmark & \checkmark \\
  ALBEF & 4.0M & \checkmark & \checkmark & \checkmark & \checkmark & \\
  VOLTA & 2.77M & & & \checkmark & & \\
  \bottomrule
\end{tabular}
\caption[Amount of pretraining data and sources of pretraining data]{Amount of pretraining data and sources of pretraining data. COCO: Microsoft COCO \citep{lin2014microsoft}, VS: Visual Genome \citep{krishna2017visual}, CC: Conceptial Captions \citep{sharma2018conceptual}, SBU: SBU Captions \citep{ordonez2011im2text}}.
\label{tab:model_data}
\end{center}
\vspace{-4mm}
\end{table}

\vspace{-4mm}
\begin{table}[!htbp]
\small
\setlength{\tabcolsep}{0.15em}
\begin{center}
\begin{tabular}{ @{} l|c @{} } 
    \toprule
     \textbf{Model} & \textbf{Num. Parameters} \\
  \midrule
  VisualBERT & 112.64 M \\ % 112637954 \\ 
  UNITER & 111.08 M \\ %& 111081986 \\
  LXMERT & 209.12 M \\ % 209124098 \\
  ALBEF & 290.34 M \\ % 290336058 \\  %580672116
  BERT & 108.31 M \\ %108311810 \\
  RoBERTa & 124.65 M \\ % 124647170\\
  VOLTA VisualBERT & 114.02 M \\
  VOLTA UNITER & 113.63 M \\
  VOLTA LXMERT & 210.50 M\\
  \bottomrule
\end{tabular}
\caption{Number of trainable parameters.}
\label{tab:model_params}
\end{center}
\vspace{-4mm}
\end{table}

\subsection{Few-shot learning experiments}
\label{sec:fewshot}

On top the zero-shot setting experiments that we conduct on all fine-tuned models, we perform few-shot learning experiments on selected fine-tuned models as shown in Table~\ref{tab:fewshot}. We select the best-performing model for the task and setting, as well as additional models to ensure that all models are tested. We adopt the same hyperparameters stated in Table~\ref{tab:hyperparams}, and perform additional fine-tuning on 200 examples of the unseen modality for 20 epochs. Typically, between 10 to 100 examples are sufficient for few-shot cross-lingual transfer on multilingual BERT \citep{zhao2021closer}. For instance, in the ``Image to Caption'' setting, we fine-tune a model already fine-tuned for 80 epochs in the image-only setting, on examples in the caption-only setting. 

\begin{table}[!h]
\centering
\small
  \begin{tabular}{l|cccc} 
  \toprule
  \textbf{Setting} & \textbf{Task} & \textbf{Models tested} & \textbf{Seed} \\
  \midrule
  \multirow{5}{*}{Image to Caption} & Spatiality & UNITER & 0 \\
   &  & LXMERT & 0 \\
   \cmidrule{2-4}
   & Quantifiers & VisualBERT & 2 \\
   &  & UNITER & 2\\
   &  & LXMERT & 2\\
  \midrule
  \multirow{5}{*}{Caption to Image} & Spatiality & UNITER & 0 \\
%   &  & VisualBERT & 0 \\
   &  & ALBEF & 2\\
   \cmidrule{2-4}
  & Quantifiers & VisualBERT & 2\\
  &  & UNITER & 2\\
  &  & ALBEF & 2 \\
  \bottomrule
  \end{tabular}
  \caption{Models tested on few-shot learning setting. (Seed: the random seed used in the initial fine-tuning of the model).} \label{tab:fewshot}
\end{table}

\section{Effect of Training Dataset Size}
\label{sec:datasetsize}
To investigate the effect of training dataset size on model performance, we fine-tune models with 50\%, 25\% and 12.5\% subsets, randomly sampled from the full training set. These experiments were conducted only on a single random seed, due to resource limitations. 

\begin{figure}[ht!]
\small
\centering
\begin{subfigure}{0.45\textwidth}
\includegraphics[width=\linewidth]{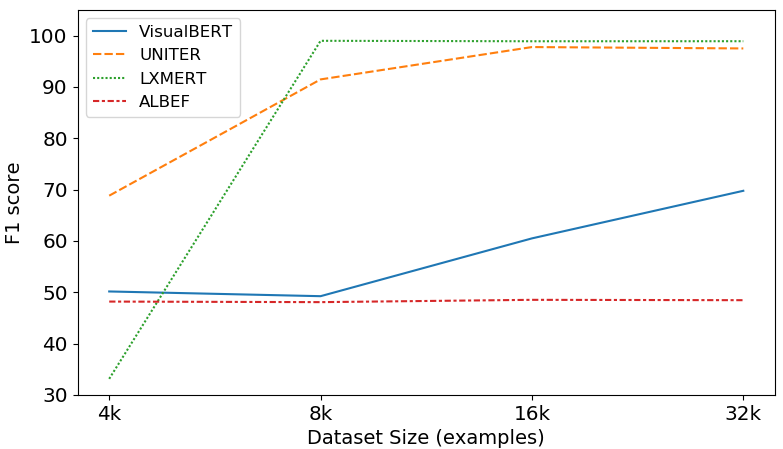}
    \caption{Image-only setting.}\label{fig:spatial_img}
\end{subfigure}

\begin{subfigure}{0.45\textwidth}
\vspace{1em}
\includegraphics[width=\linewidth]{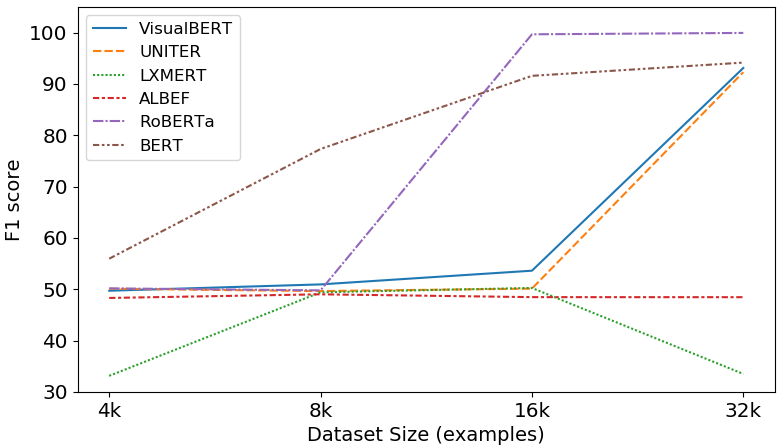}
    \caption{Caption-only setting.}\label{fig:spatial_cap}
\end{subfigure}

\begin{subfigure}{0.45\textwidth}
\vspace{1em}
\includegraphics[width=\linewidth]{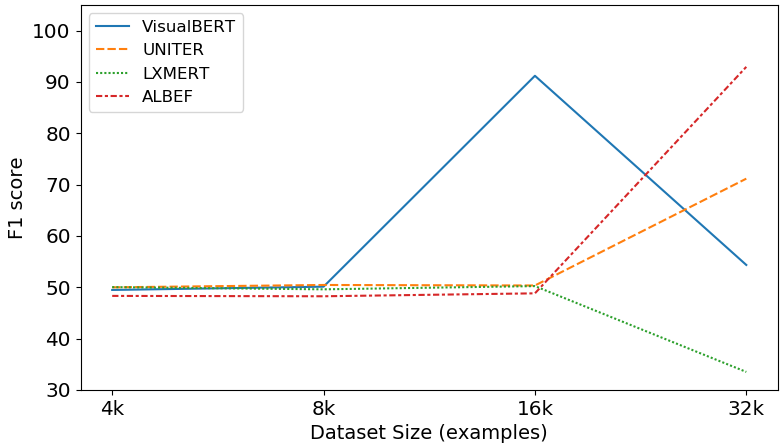}
    \caption{Image+caption setting.}\label{fig:spatial_img_cap}
\end{subfigure}
\caption{Performance for the spatiality task on the OOD test set in (a) image-only, (b) caption-only and (c) image+caption settings; models trained on increasing subsets with one random seed.}
\vspace{-0.2em}
\label{fig:spatial_size}
\end{figure}

\paragraph{Spatiality.} On the spatiality task (Figure~\ref{fig:spatial_size}), in the image-only setting, LXMERT requires the least amount of data to achieve convergence (4k examples), followed by UNITER (8k examples), and VisualBERT (16k examples). ALBEF fails to converge, even on the full 32k dataset. There is a clear advantage in performance of LXMERT and UNITER over VisualBERT and ALBEF. 

In the caption-only setting, the text-only model RoBERTa performs the best, achieving F$_1$s close to 100 with 16k training examples. While BERT, UNITER and VisualBERT achieve similar results with 32k training examples, the V+L models are outperformed by BERT when trained on a smaller dataset. 

\begin{figure}[ht!]
\small
\centering
\begin{subfigure}{0.48\textwidth}
\includegraphics[width=\linewidth]{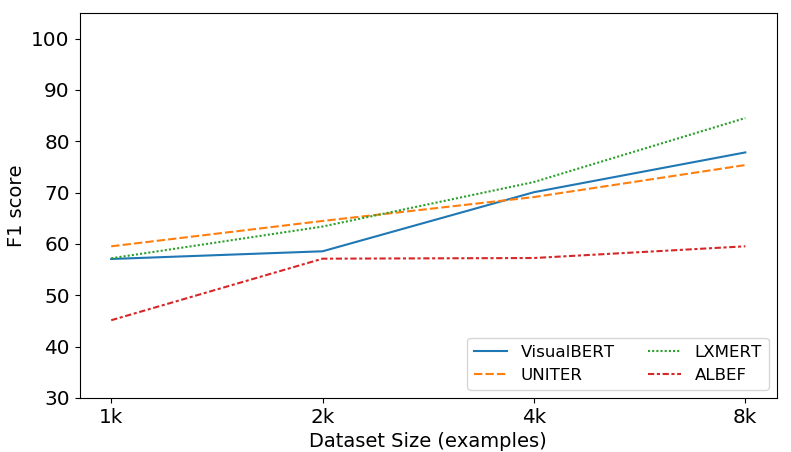}
    \caption{Image-only setting.}\label{fig:cardinality_img}
\end{subfigure}
\hspace{\fill}
\begin{subfigure}{0.48\textwidth}\vspace{1em}
\includegraphics[width=\linewidth]{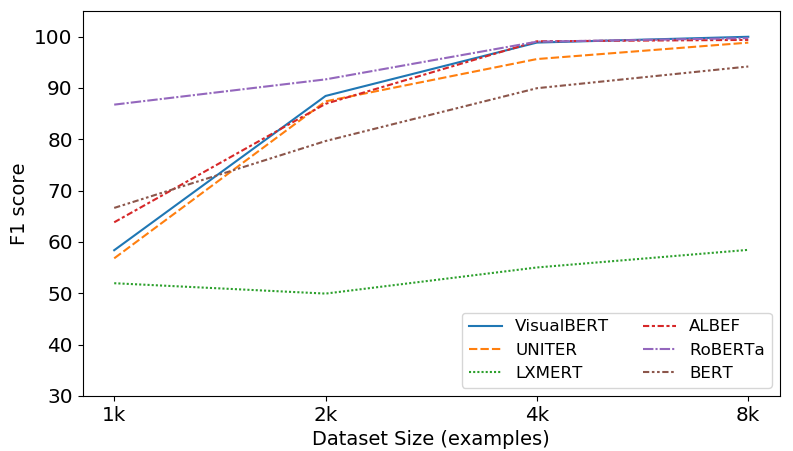}
    \caption{Caption-only setting.}\label{fig:cardinality_cap}
\end{subfigure}

\begin{subfigure}{0.48\textwidth}\vspace{1em}
\includegraphics[width=\linewidth]{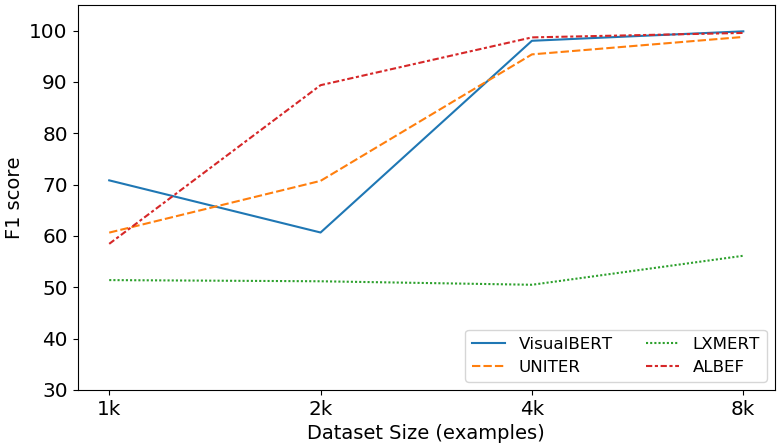}
    \caption{Image+caption setting.}\label{fig:cardinality_img_cap}
\end{subfigure}

\caption{Performance for the cardinality task on the OOD test set in (a) image-only, (b) caption-only and (c) image+caption settings; models trained on increasing subsets with one random seed.}
\vspace{-0.2em}
\label{fig:card_size}
\end{figure}

\paragraph{Cardinality.} Compared to the spatiality task, more models converged on the cardinality task (Figure~\ref{fig:card_size}) with a smaller amount of data (8k examples). In the image-only setting, LXMERT remains the best performing model on the full 8k dataset, but the performance of VisualBERT, UNITER and LXMERT are relatively similar. In the caption-only setting, when trained on a smaller dataset with 1k examples, RoBERTA achieves an F$_1$ of 86.76, while all other models achieve an F$_1$ between 55 and 67, except for LXMERT, which fails to achieve non-random performance. In the image+caption setting, ALBEF seems to be most data efficient, performing the best on a 2k training dataset.

\begin{figure}[ht!]
\small
\centering
\begin{subfigure}{0.48\textwidth}
\includegraphics[width=\linewidth]{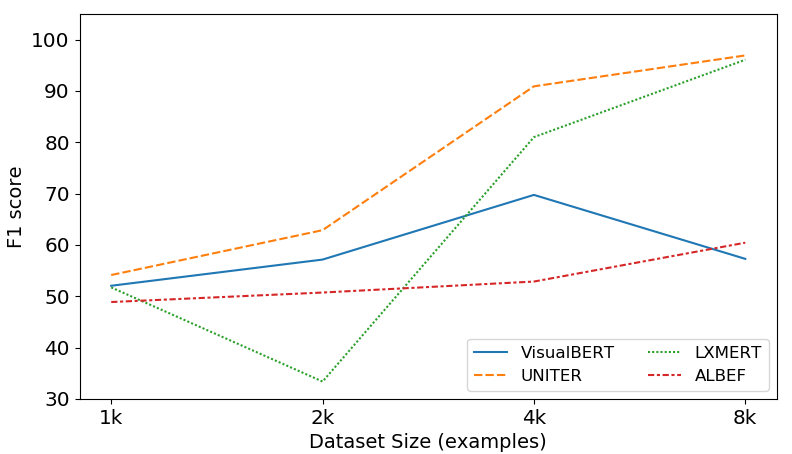}
    \caption{Image-only setting.}\label{fig:quantifier_img}
\end{subfigure}
\hspace{\fill}
\begin{subfigure}{0.48\textwidth}\vspace{1em}
\includegraphics[width=\linewidth]{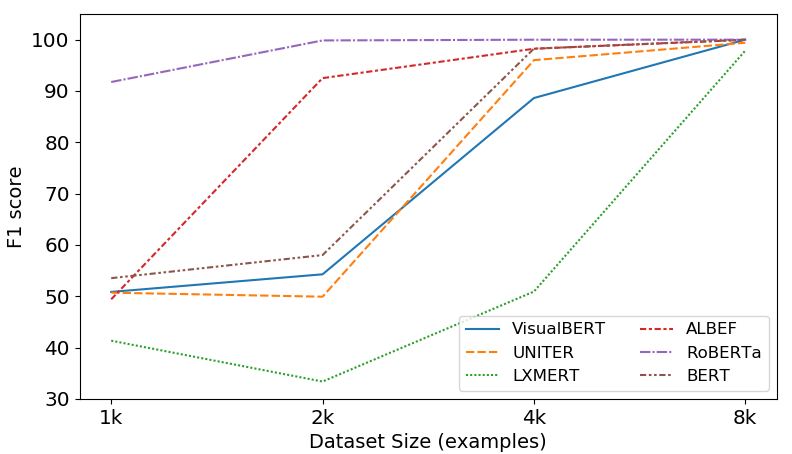}
    \caption{Caption-only setting.}\label{fig:quantifier_cap}
\end{subfigure}

\begin{subfigure}{0.48\textwidth}\vspace{1em}
\includegraphics[width=\linewidth]{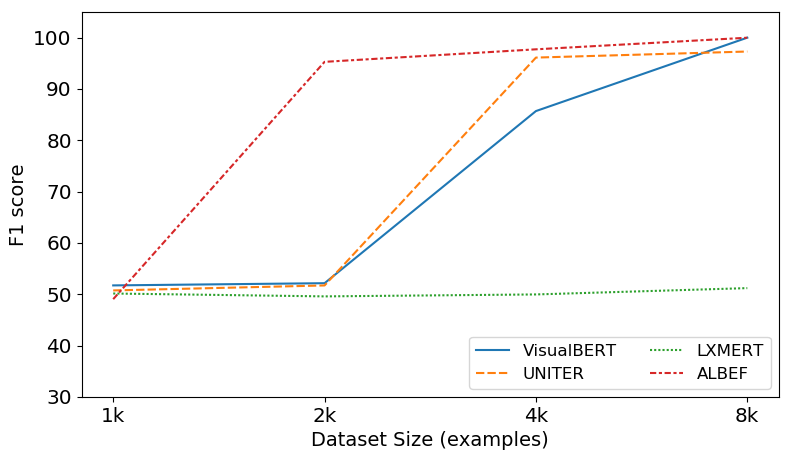}
    \caption{Image+caption setting.}\label{fig:quantifier_img_cap}
\end{subfigure}

\caption{Performance for the quantifiers task on the OOD test set in (a) image-only, (b) caption-only and (c) image+caption settings; models trained on increasing subsets with one random seed.}
\vspace{-0.2em}
\label{fig:quantifier_size}
\end{figure}

\paragraph{Quantifiers.} On the quantifiers task (Figure~\ref{fig:quantifier_size}), in the image-only setting, we observe a similar overall trend to the spatiality task, where UNITER and LXMERT significantly outperform the other models. In the caption-only setting, RoBERTa significantly outperforms BERT with limited data (1k examples), and the performance of UNITER and VisualBERT resembles that of BERT. The superior performance of ALBEF on a smaller 2k dataset in both caption-only and image+caption settings is observed again, as on the cardinality task.

\begin{figure}[ht!]
\small
\centering
\begin{subfigure}{0.48\textwidth}
\includegraphics[width=\linewidth]{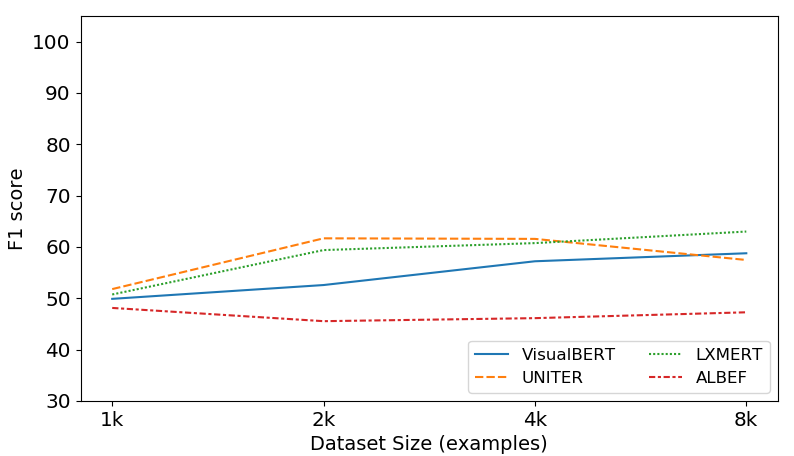}
    \caption{Image-only setting.}\label{fig:compare_img}
\end{subfigure}
\hspace{\fill}
\begin{subfigure}{0.48\textwidth}
\vspace{1em}
\includegraphics[width=\linewidth]{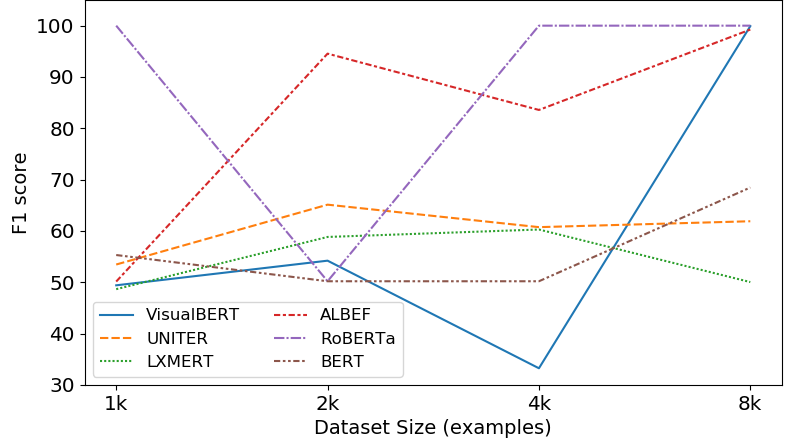}
    \caption{Caption-only setting.}\label{fig:compare_cap}
\end{subfigure}

\begin{subfigure}{0.48\textwidth}
\vspace{1em}
\includegraphics[width=\linewidth]{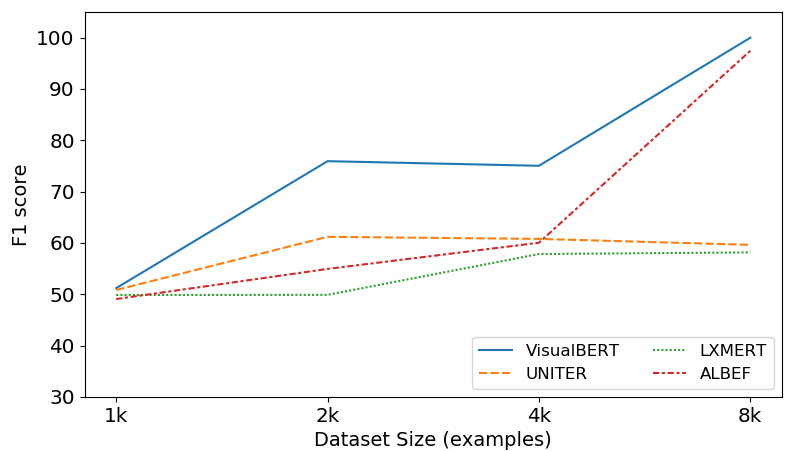}
    \caption{Image+caption setting.}\label{fig:compare_img_cap}
\end{subfigure}
\caption{Performance for the numerical comparison task on the OOD test set in (a) image-only, (b) caption-only and (c) image+caption settings; models trained on increasing subsets with one random seed.}
\vspace{-0.2em}
\label{fig:compare_size}
\end{figure}

\paragraph{Numerical Comparison.} On the numerical comparison task (Figure~\ref{fig:compare_size}), all models exhibit relatively poor performance on the OOD dataset in the image-only setting, revealing an inability to generalise to unseen number pairs. In the caption-only and image+caption settings, the superior performance of RoBERTa over BERT, and VisualBERT and ALBEF over the other V+L models can be observed, as similarly described for other tasks. The performance of VisualBERT and UNITER is clearly distinguished. 

\section{Training Duration}
\begin{figure}[h!]
\small
\centering
\begin{subfigure}{0.48\textwidth}
\includegraphics[width=\textwidth]{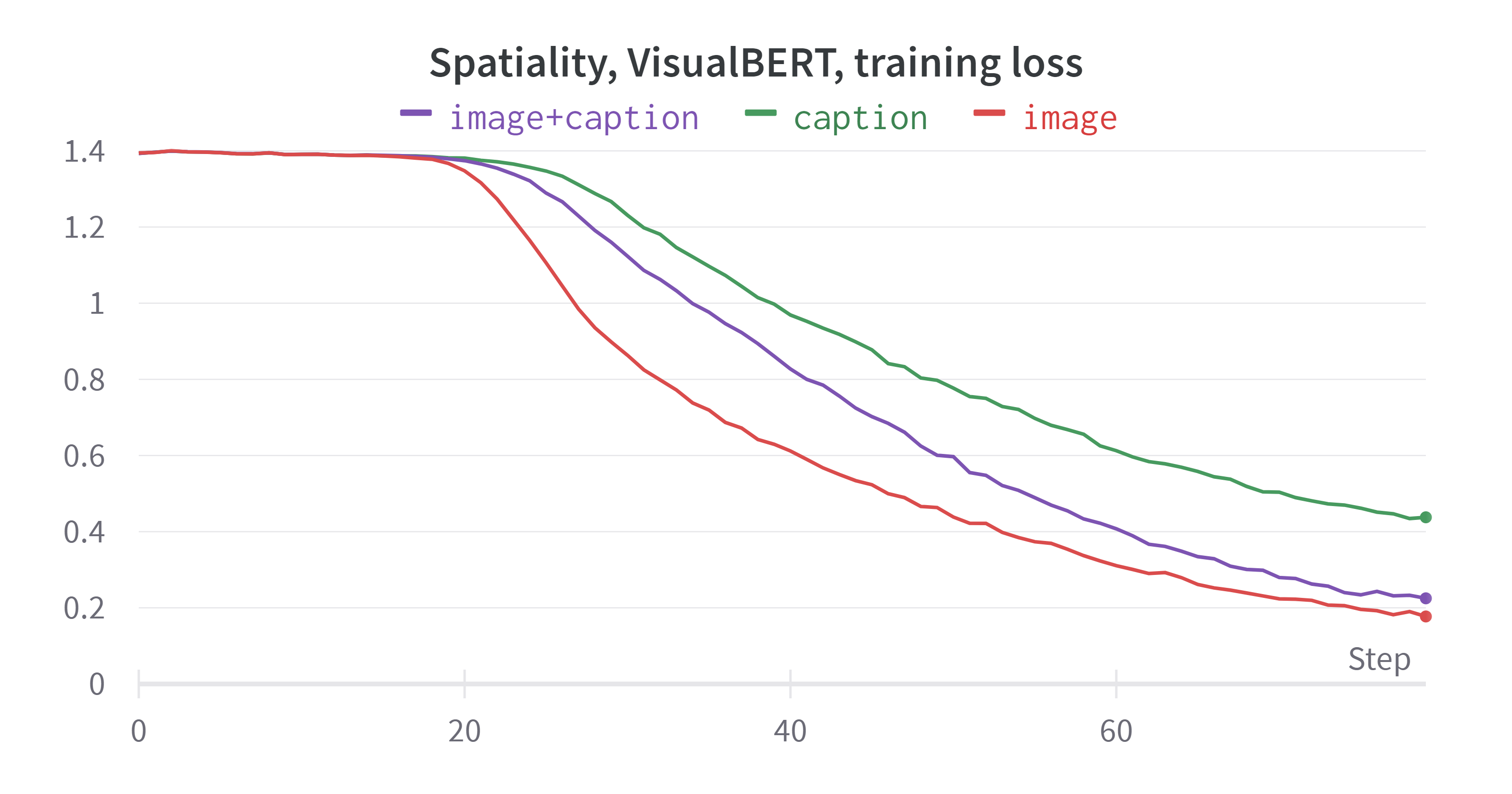}
\caption{VisualBERT (seed=1)}
\end{subfigure}
\hspace{\fill}
\begin{subfigure}{0.48\textwidth}
\includegraphics[width=\textwidth]{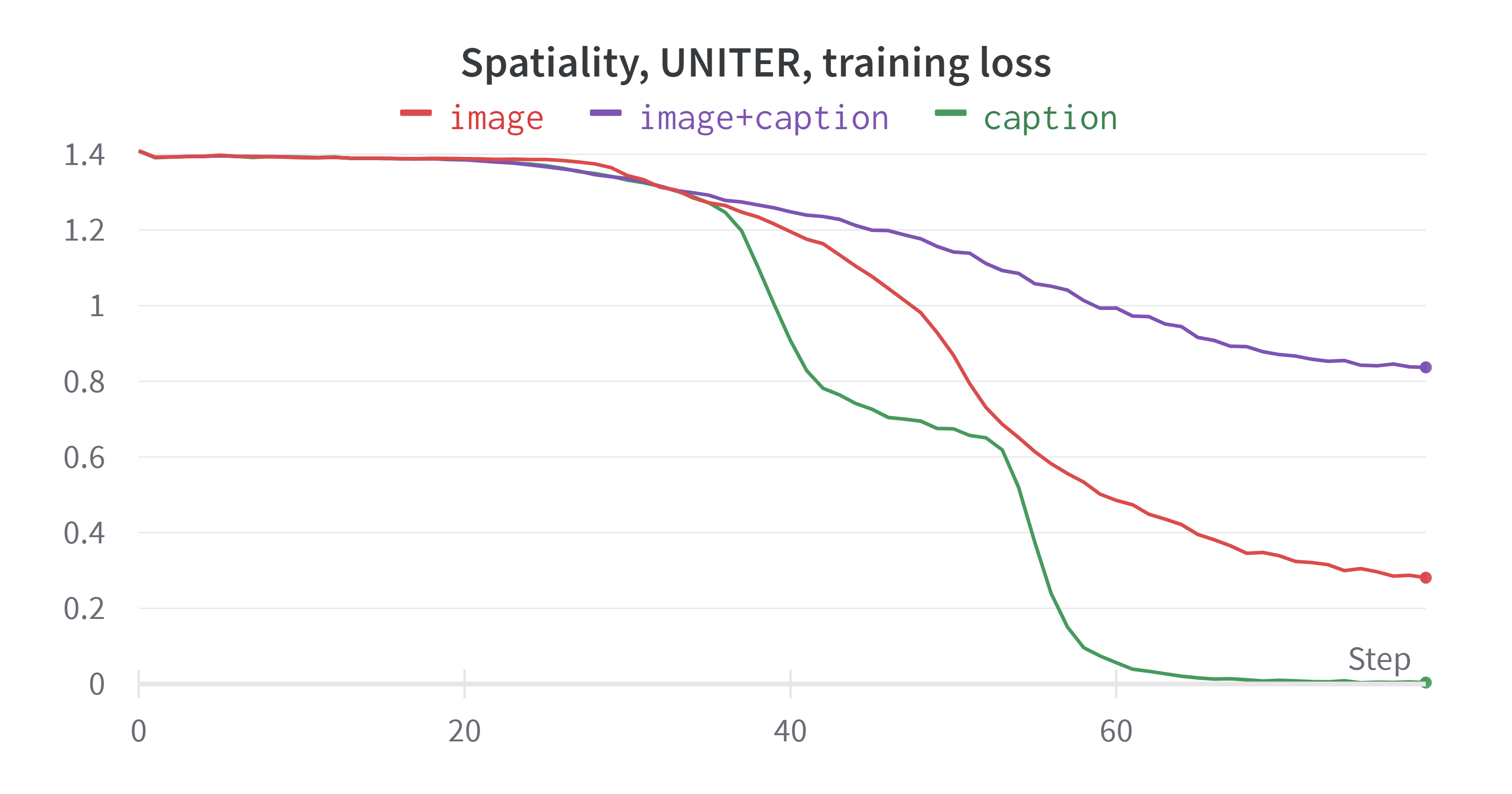}
\caption{UNITER (seed=0)}
\end{subfigure}
\hspace{\fill}
\begin{subfigure}{0.48\textwidth}\vspace{1em}
\includegraphics[width=\textwidth]{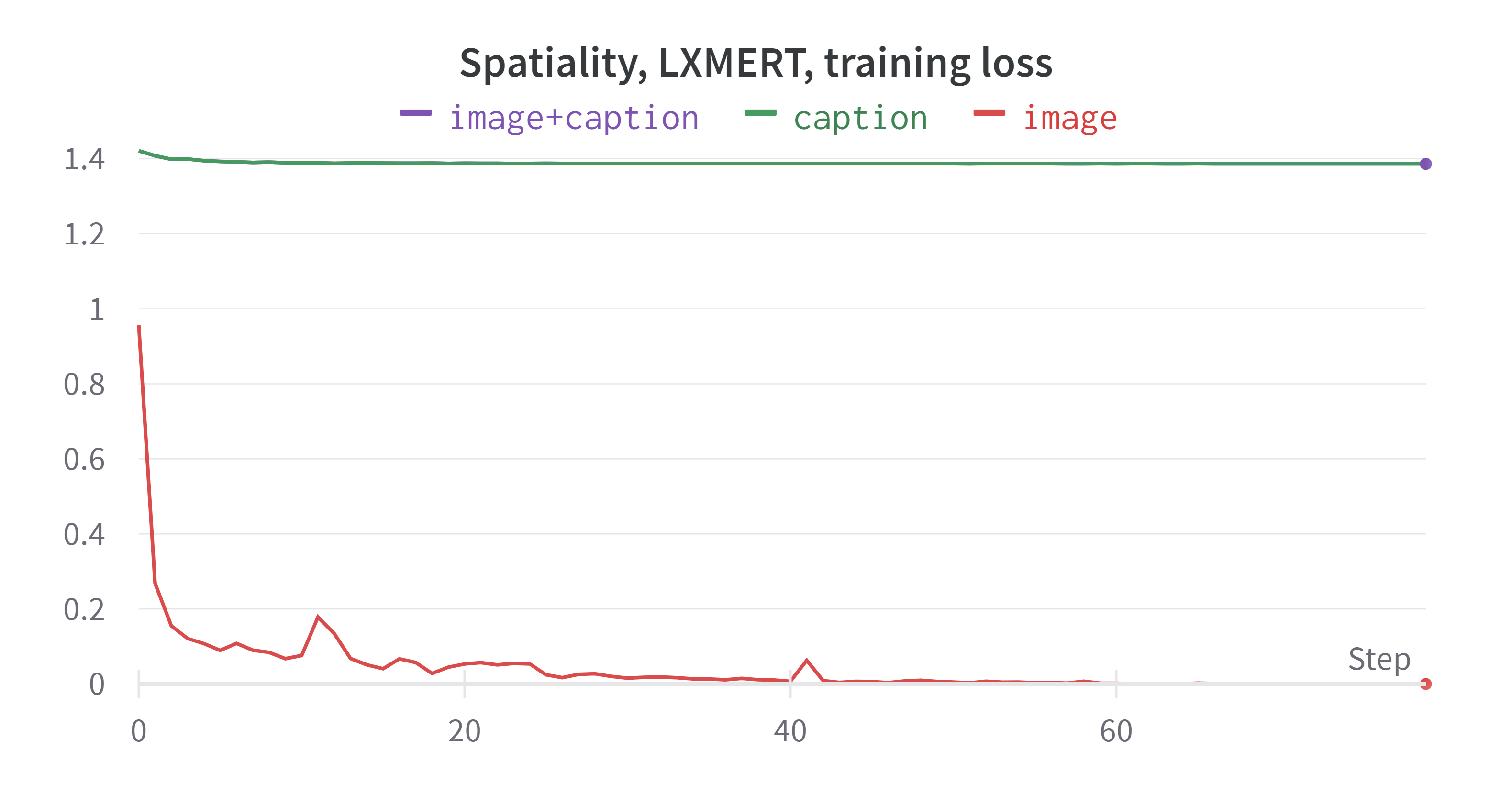}
\caption{LXMERT (seed=1)}
\end{subfigure}
\hspace{\fill}
\begin{subfigure}{0.48\textwidth}\vspace{1em}
\includegraphics[width=\textwidth]{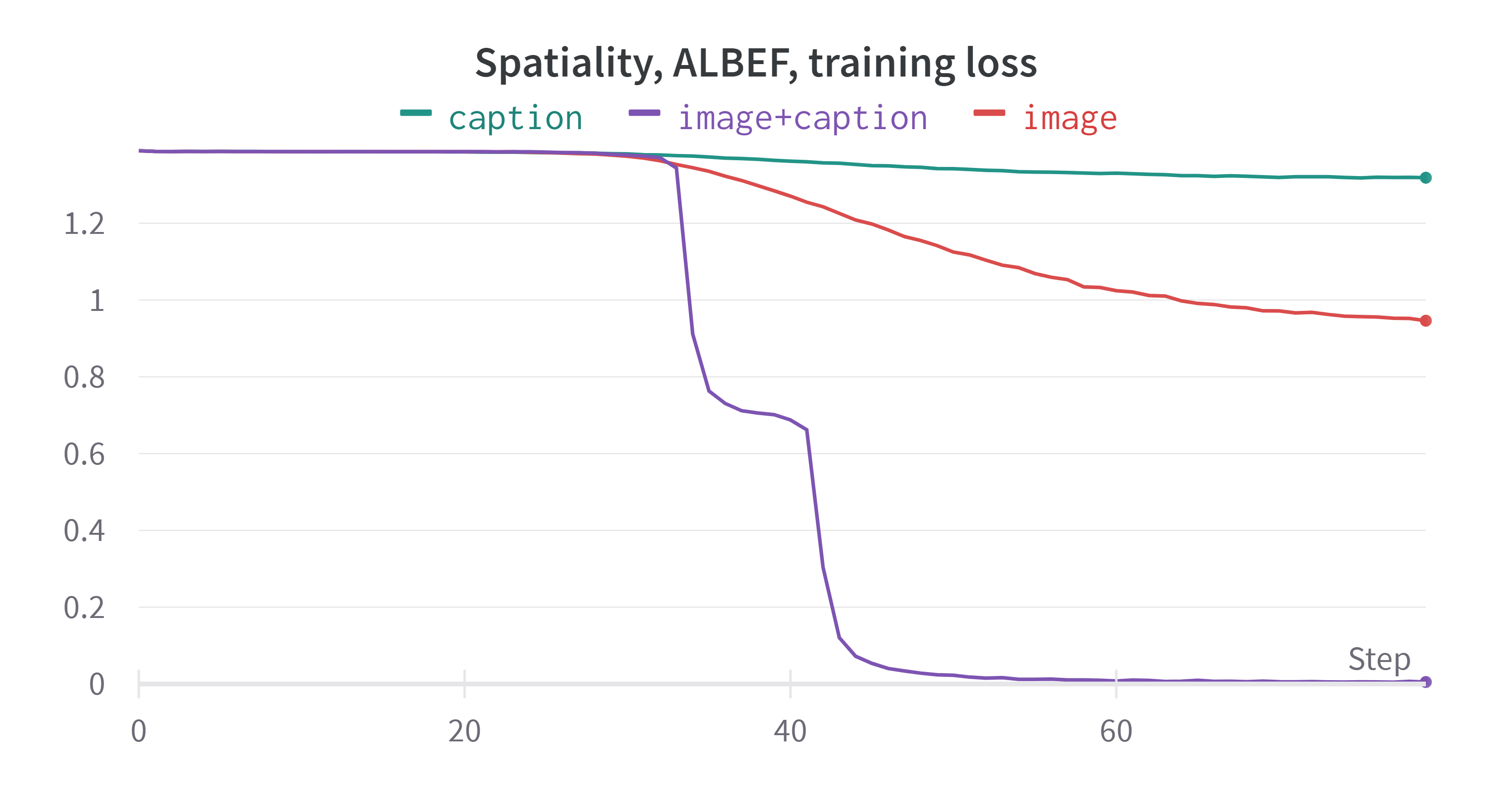}
\caption{ALBEF (seed=0)}
\end{subfigure}
\caption{Training loss curves on spatiality task.}
\label{fig:loss1}
\vspace{-4mm}
\end{figure}

We discuss selected runs indicative of the contrast in the number of training steps required by models. From the training loss curves on a single run of training on the spatiality tasks, we can observe a substantial degree of variation in the amount of time each model takes to converge.

The training loss curves in Figure~\ref{fig:loss1} are on the whole indicative of the significant amount of time required by all models except LXMERT to learn the spatiality task. With the notable exception of LXMERT, many models have yet to fully converge by 60 epochs. LXMERT takes fewer than 5 epochs to achieve a training loss close to 0, while completely failing to converge on the caption and image+caption settings. Possible reasons for the faster convergence of LXMERT on the spatiality task include the large size of its pretraining data, and the fact that it was additionally pretrained on a VQA task, unlike the other models. VisualBERT has yet to completely converge after 80 epochs. ALBEF in the image+caption setting and UNITER in the caption setting are representative of successful convergence, which is achieved only after 40 and 60 epochs, respectively.

\begin{figure*}[ht!]
\small
\centering
\begin{subfigure}{0.48\textwidth}
\includegraphics[width=\textwidth]{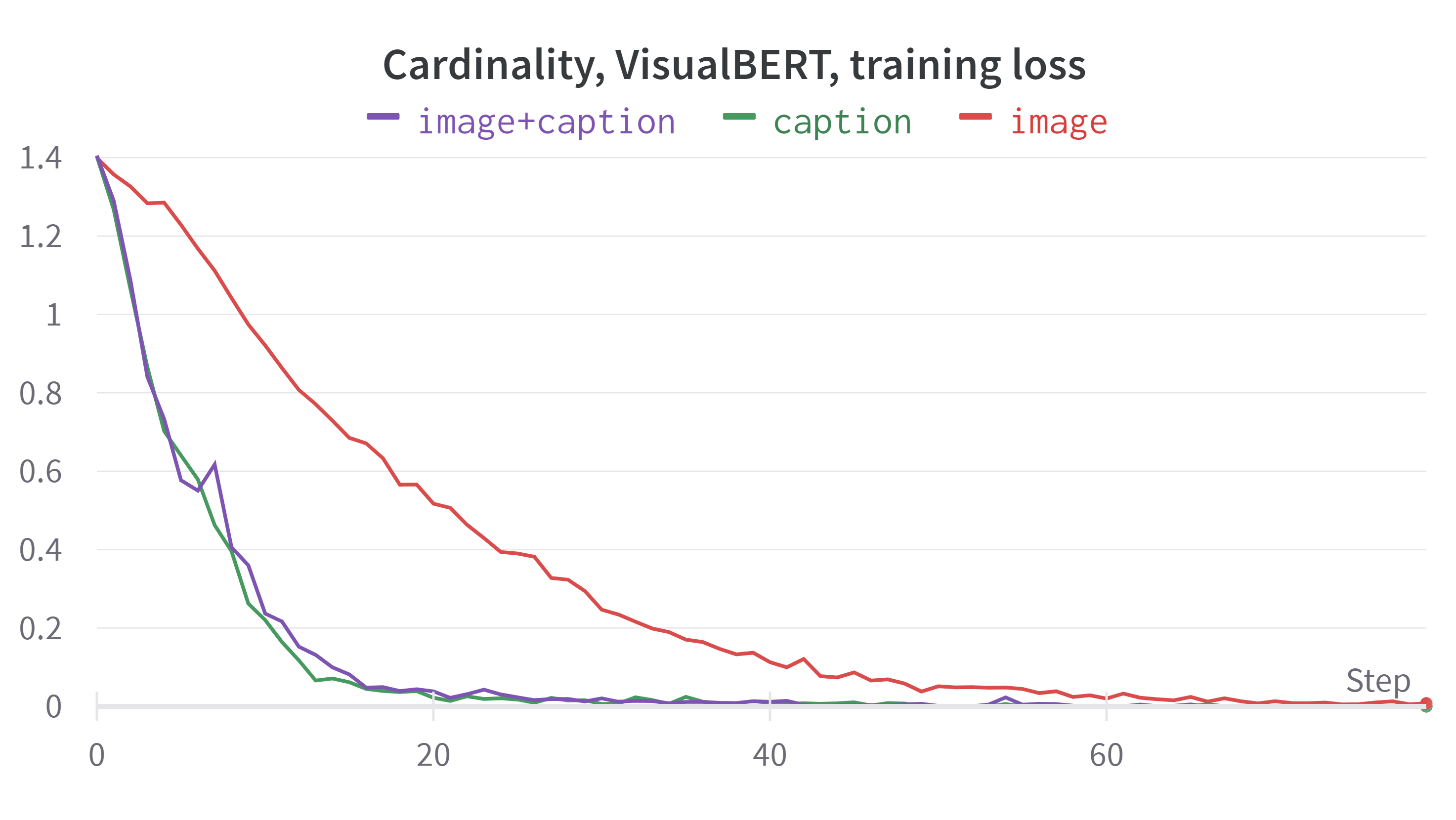}
\caption{VisualBERT (seed=1)}
\end{subfigure}
\hspace{\fill}
\begin{subfigure}{0.48\textwidth}
\includegraphics[width=\textwidth]{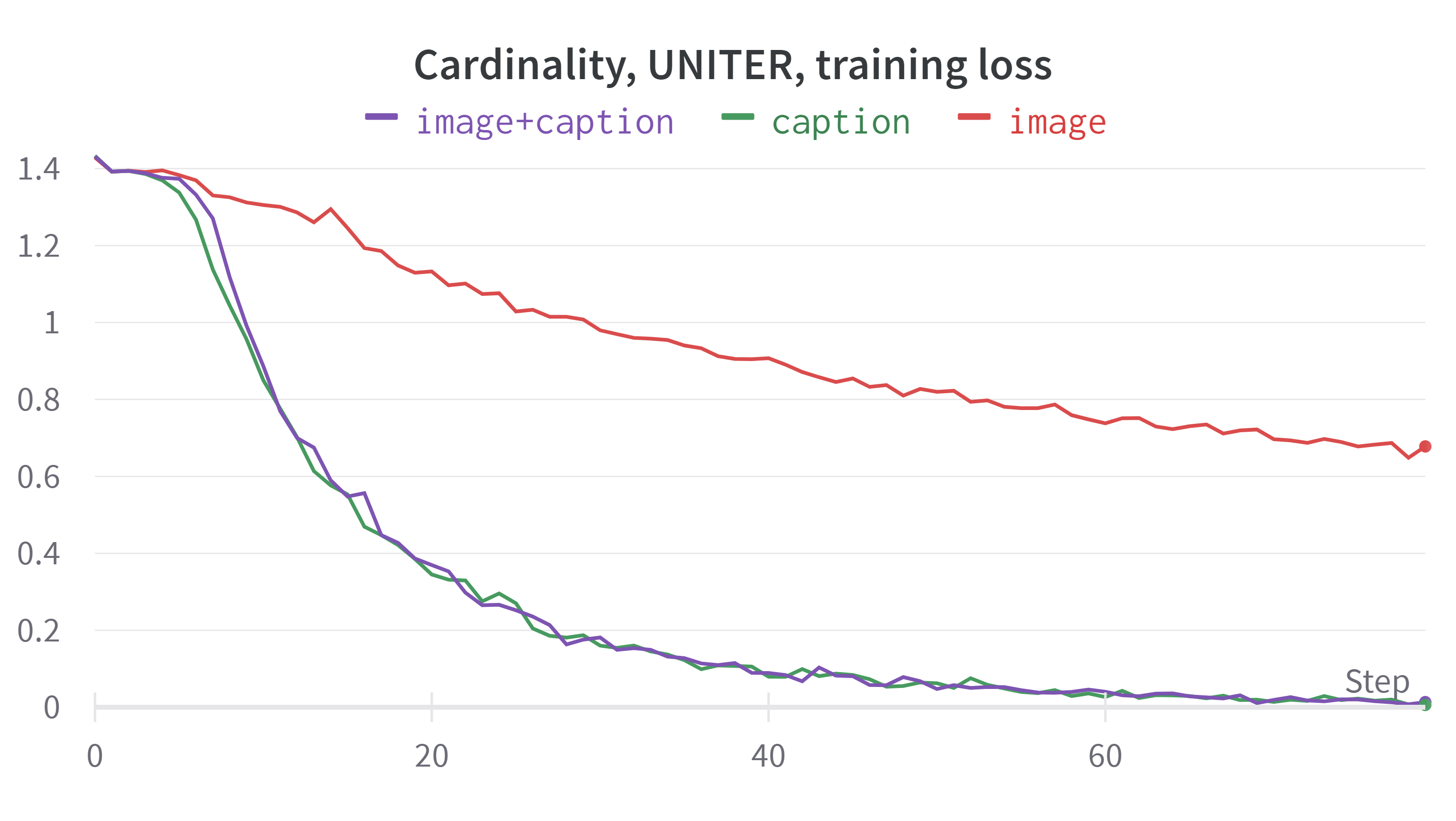}
\caption{UNITER (seed=0)}
\end{subfigure}
\hspace{\fill}
\begin{subfigure}{0.48\textwidth}\vspace{1em}
\includegraphics[width=\textwidth]{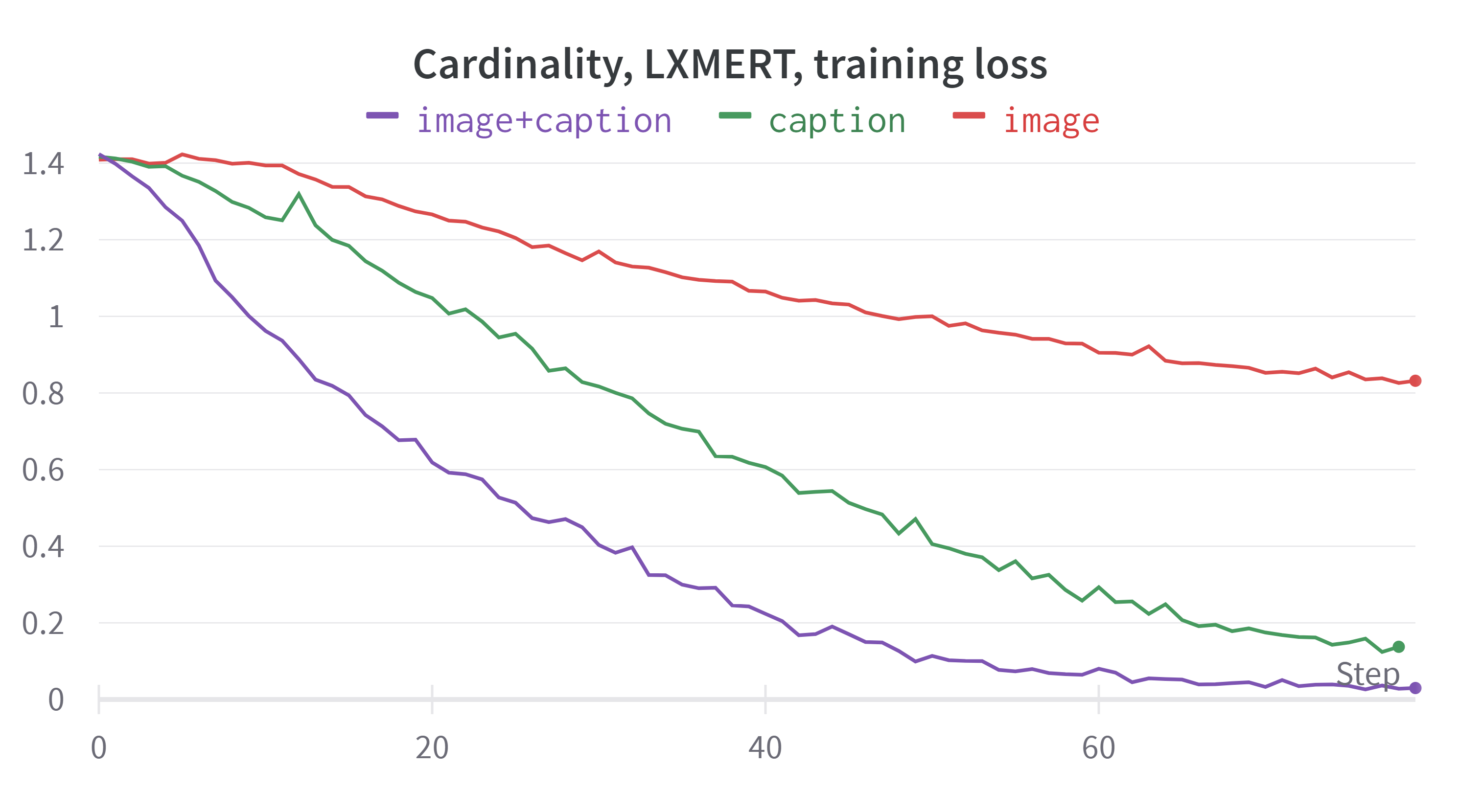}
\caption{LXMERT (seed=1)}
\end{subfigure}
\hspace{\fill}
\begin{subfigure}{0.48\textwidth}\vspace{1em}
\includegraphics[width=\textwidth]{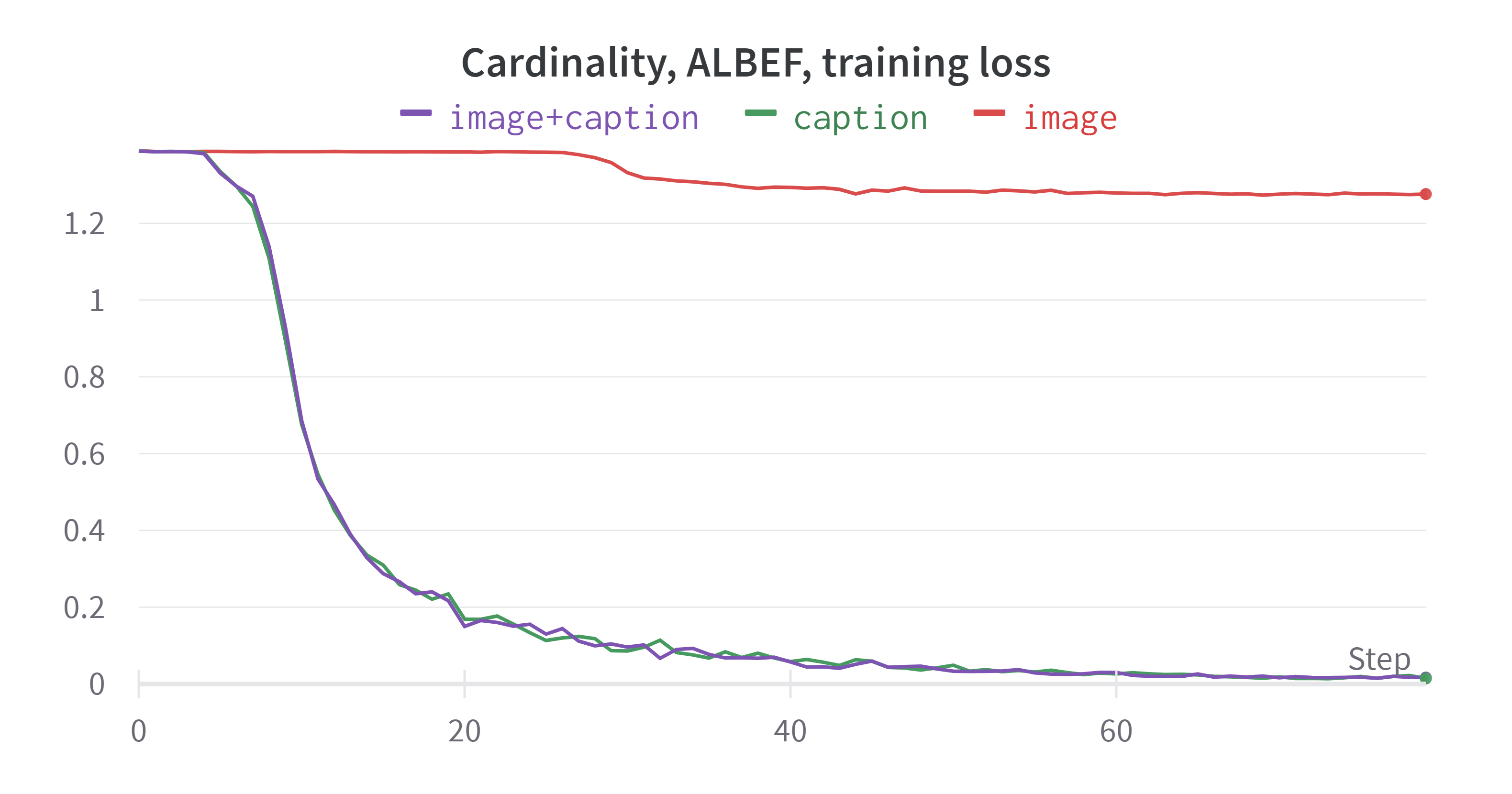}
\caption{ALBEF (seed=0)}
\end{subfigure}
\caption{Training loss curves on cardinality task.}
\label{loss2}
\end{figure*}

The variation between models is much less significant on the cardinality dataset, as seen in Figure~\ref{loss2}. The training loss curves between the caption-only and image+caption settings are often similar, and also steeper compared to the image-only loss curves, which is reflective of the textual bias of all models towards the caption when both modalities are presented in the image+caption setting. In the image-only setting, most models have yet to fully converge within 80 epochs, in this case, with the exception of VisualBERT. It is notable that the marked advantage of LXMERT in terms of efficiency of training time observed in the spatiality task does not extend to other tasks.
\label{sec:training_efficiency}

\FloatBarrier
\section{VOLTA Experiments}
\label{sec:volta}
From Table \ref{tab:volta}, we observe that while the performance of all three \textsc{Volta} models are similar on the 32k spatiality dataset, \textsc{Volta} UNITER significantly outperforms both \textsc{Volta} LXMERT and \textsc{Volta} VisualBERT on the 8k spatiality dataset. This is in contrast to results on the original models where LXMERT exhibits a clear advantage over the other models and is able to achieve an F$_1$ score close to 100 on the 8k dataset. 

\begin{table}[h!]
\tablesize
\setlength{\tabcolsep}{0.5em}
\begin{center}
\begin{tabular}{ @{} c|c|c|c|c|c|c @{} } 
%   \toprule
%   \multicolumn{9}{c}{\small \textbf{Spatiality}}\\
    \midrule
     \textbf{Dataset} &  \textbf{Model} 
   & \multicolumn{2}{c|}{\textbf{OOD}}
   & \multicolumn{2}{c|}{\textbf{InD}}
   & \textbf{\# Seeds} \\
     \textbf{Size} &
   & \textbf{Mean} & \textbf{Stdev.}
   & \textbf{Mean} & \textbf{Stdev.} 
   & \\
  \midrule
  \multirow{6}{*}{32k} & \textsc{Volta} LXMERT & 99.30 & - & 99.67 & - & 1\\
  &  \textsc{Volta} UNITER & 99.67 & - & 99.88 & - & 1\\
  & \textsc{Volta} VisualBERT & 98.31 & - & 98.92 & - & 1 \\
  
  & LXMERT  & 99.46 & 0.47 & 99.84 & 0.13 & 3\\
  & UNITER & 89.67 & 14.43 & 88.71 & 14.58 & 3\\
  & VisualBERT & 65.94 & 6.63 & 66.96 & 6.05 & 3\\
  \midrule
  
  \multirow{6}{*}{8k} & \textsc{Volta} LXMERT & 62.28 & 3.16	& 64.17  & 3.73 & 3\\
  & \textsc{Volta} UNITER & 94.46  & 1.19 & 93.18 & 1.50 & 3\\
  & \textsc{Volta} VisualBERT & 60.73 & 2.39 & 60.69 & 2.90 & 3 \\

  & LXMERT & 99.02 & - &  99.00 & - & 1\\
  & UNITER & 88.33 & - & 91.48 & - & 1\\
  & VisualBERT & 49.60 & - & 49.25 & - & 1\\
  \bottomrule
\end{tabular}
\caption[Mean F$_1$ scores of \textsc{Volta} and original models trained on spatiality dataset in image-only setting.]{Mean F$_1$ scores of \textsc{Volta} and original models trained on spatiality dataset in the image-only setting. Standard deviation is shown where applicable.}
\label{tab:volta}
\end{center}
\vspace{-4mm}
\end{table}

Similar findings are observed on the cardinality dataset (Table \ref{tab:volta2}), where  \textsc{Volta} LXMERT is outperformed by both \textsc{Volta} UNITER and \textsc{Volta} VisualBERT. The \textsc{Volta} LXMERT differs from the original LXMERT in the amount of pretraining data, as well as the use of VQA examples and a downstream VQA objective during pretraining. It can be concluded that LXMERT loses its advantage in the image-only setting when it is pretrained in the same manner as other models. 

\begin{table}[h!]
\tablesize
\begin{center}
\begin{tabular}{ @{} c|c|c|c|c|c @{} } 
%   \toprule
%   \multicolumn{9}{c}{\small \textbf{Spatiality}}\\
    \midrule
    \textbf{Model} 
   & \multicolumn{2}{c|}{\textbf{OOD}}
   & \multicolumn{2}{c|}{\textbf{InD}}
   & \textbf{\# Seeds} \\
   & \textbf{Mean} & \textbf{Stdev.}
   & \textbf{Mean} & \textbf{Stdev.} 
   & \\
  \midrule
  
  \textsc{Volta} LXMERT & 73.38 & 0.56 & 79.30 & 0.39 & 3\\ 
  \textsc{Volta} UNITER & 85.91 & 5.13 & 87.40 & 3.36 & 3\\
  \textsc{Volta} VisualBERT & 81.67 & 4.46 & 83.90 & 1.66 & 3\\
  
  LXMERT & 82.90 & 3.86 & 84.56 & 2.71 & 3\\
  UNITER & 77.12 & 5.04 & 77.03 & 1.44 & 3\\
  VisualBERT & 77.41 & 0.61 & 76.30 & 0.46 & 3\\
  \bottomrule
\end{tabular}
\caption[Mean F$_1$ scores of \textsc{Volta} and original models trained on 8k cardinality dataset in image-only setting.]{F$_1$ scores of \textsc{Volta} and original models trained on cardinality dataset in the image-only setting. Standard deviation is shown where applicable.} 
\label{tab:volta2}
\end{center}
\vspace{-4mm}
\end{table}

\begin{figure}[h]
\small
\centering
\begin{subfigure}{0.48\textwidth}
\includegraphics[width=\textwidth]{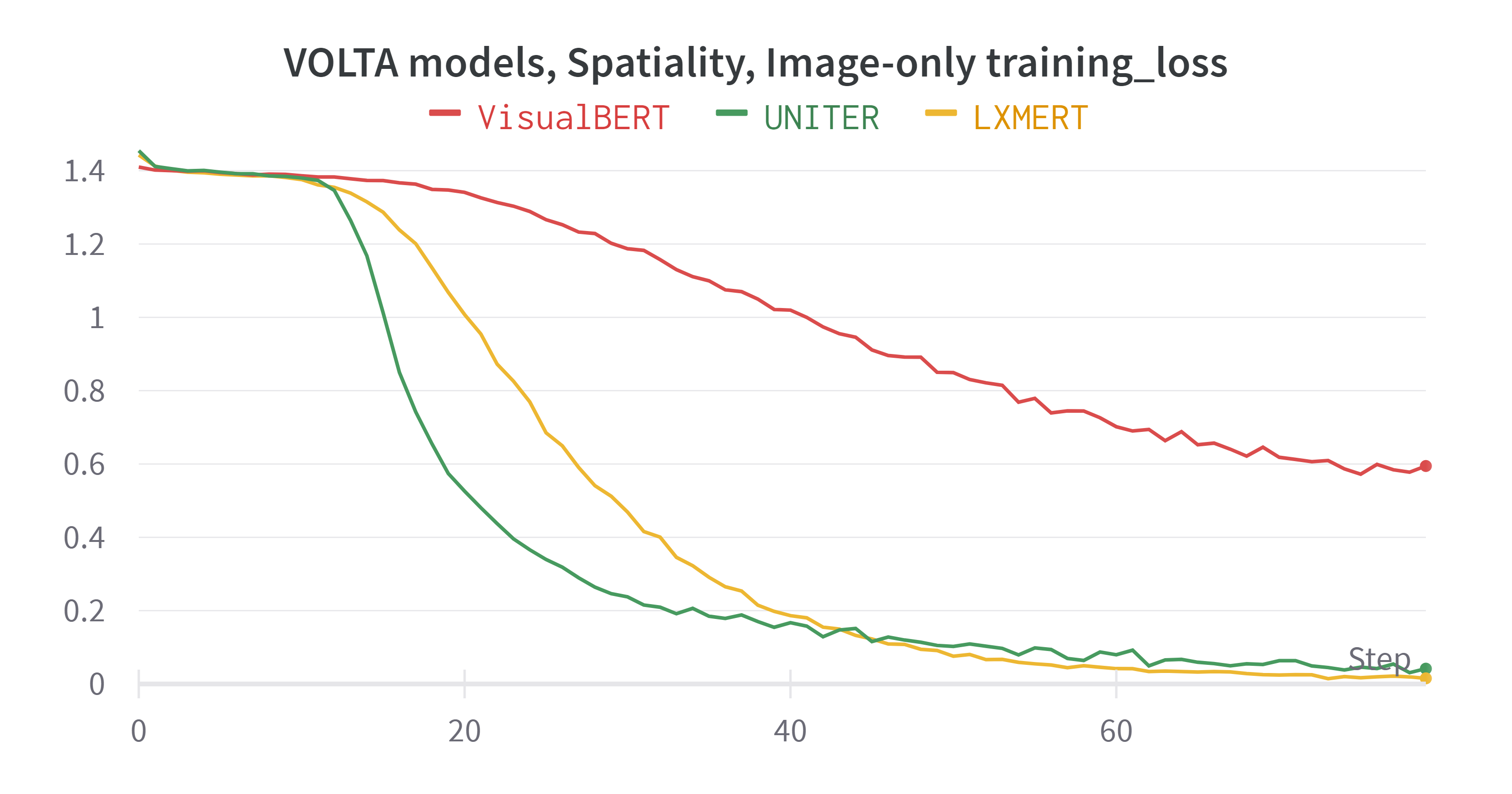}
\caption{\textsc{Volta} models}
\end{subfigure}
\hspace{\fill}
\begin{subfigure}{0.48\textwidth}
\includegraphics[width=\textwidth]{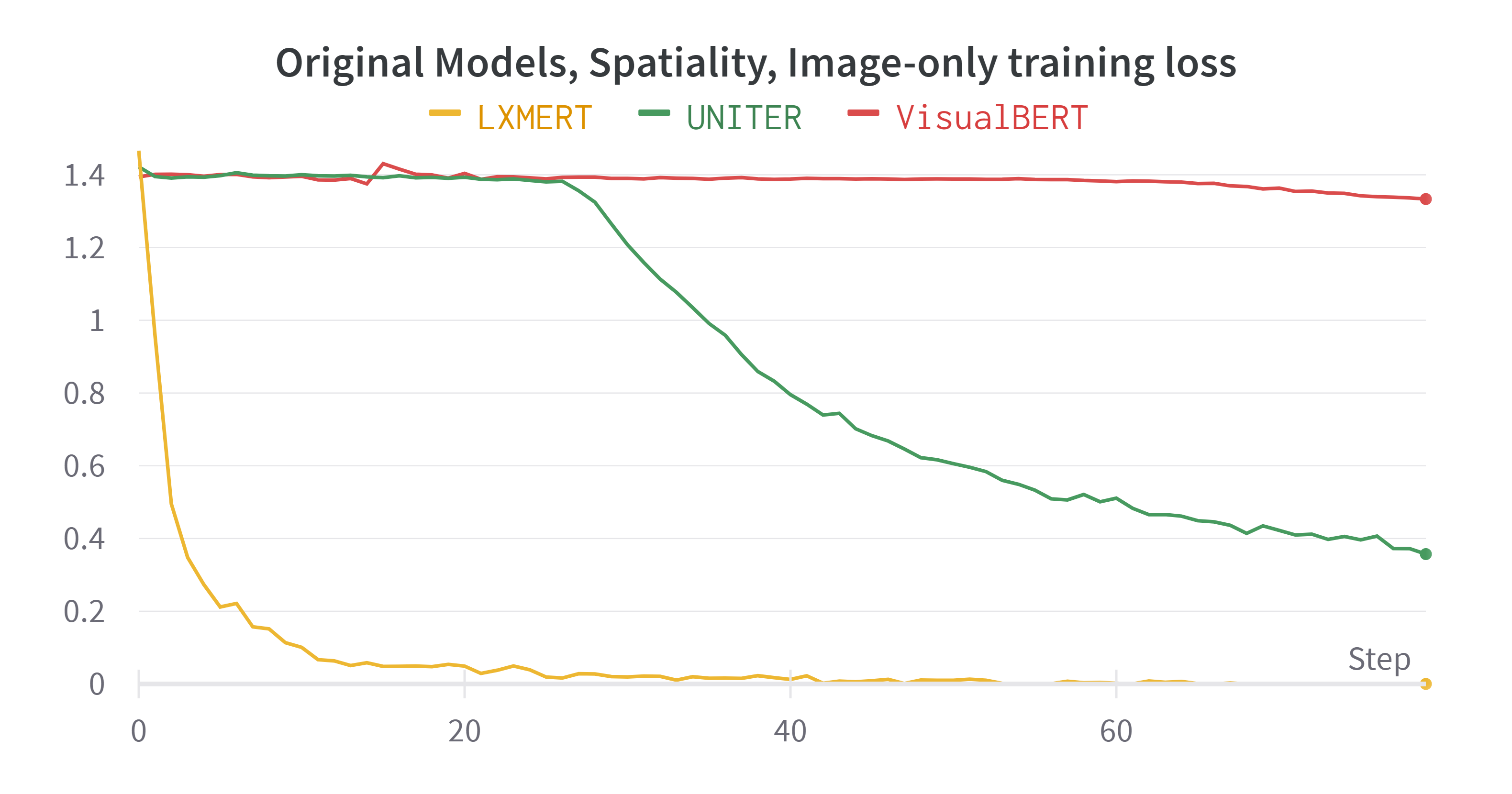}
\caption{Original models}
\end{subfigure}
\hspace{\fill}
\caption{Training loss curves on 8k spatiality dataset in the image-only setting.}
\label{fig:volta-loss}
\end{figure}

The difference between \textsc{Volta} LXMERT and the original LXMERT can also be clearly observed in the training loss curves in Figure~\ref{fig:volta-loss}, where the original LXMERT is significantly more efficient in terms of number of epochs required for training.

Furthermore, it is notable that the performance of VisualBERT on the full dataset is significantly improved over the original model. This is likely due to the fact that the \textsc{Volta} VisualBERT was pretrained with more data than the original VisualBERT. An additional finding is that \textsc{Volta} UNITER outperforms \textsc{Volta} VisualBERT and LXMERT. 

\begin{table}[ht!]
\tablesize
\begin{center}
\begin{tabular}{ @{} c|c|c|c|c|c @{} } 
%   \toprule
%   \multicolumn{9}{c}{\small \textbf{Spatiality}}\\
    \midrule
     \textbf{Model} 
   & \multicolumn{2}{c|}{\textbf{OOD}}
   & \multicolumn{2}{c|}{\textbf{InD}}
   & \textbf{\# Seeds} \\
   
   & \textbf{Mean} & \textbf{Stdev.}
   & \textbf{Mean} & \textbf{Stdev.} 
   & \\
  \midrule
  
  \textsc{Volta} LXMERT & 97.87 & 0.85 & 99.44 & 0.03 & 3 \\ 
  \textsc{Volta} UNITER & 98.73 & 1.25 & 99.60 & 0.11 & 3 \\
  \textsc{Volta} VisualBERT & 99.40 & 0.03 & 98.50 & 0.11 &  3\\
  
  LXMERT & 60.02 & 4.76 & 78.90 & 6.83 & 3\\
  UNITER & 98.96 & 0.30 & 97.14 & 0.83 & 3\\
  VisualBERT & 99.99 & 0.01 & 99.91 & 0.06 & 3\\
  \bottomrule
\end{tabular}
\caption[Mean F$_1$ scores of \textsc{Volta} and original models trained on cardinality dataset in caption-only setting.]{Mean F$_1$ scores of \textsc{Volta} and original models trained on the 8k cardinality dataset in the caption-only setting. Standard deviation is shown where applicable.} 
\label{tab:volta3}
\end{center}
\vspace{-4mm}
\end{table}

Finally, the \textsc{Volta} models were also used to investigate the comparatively poor performance of the original LXMERT on the caption-only and image+caption settings. As seen in Table \ref{tab:volta3}, all three \textsc{Volta} models achieve similar results on the 8k cardinality dataset. Given that all \textsc{Volta} models were initialised with BERT weights, this lends support to the idea that the poor performance of LXMERT on textual input is due to the lack of initialisation with BERT weights \citep{tan2019lxmert}.

\clearpage
\begin{table*}[ht!]
\fontsize{8}{12}\selectfont
\begin{center}
\begin{tabular}{ @{} c|c|ccc|ccc|c @{} } 
%   \toprule
%   \multicolumn{9}{c}{\small \textbf{Spatiality}}\\
    \midrule
     \textbf{\uline{Task}} &  \textbf{\uline{Model}} 
   & \multicolumn{3}{c|}{\textbf{\uline{OOD}}} 
   & \multicolumn{3}{c|}{\textbf{\uline{InD}}} & \textbf{\uline{\#sd}} \\
   
& & \textbf{Image} & \textbf{Caption} & \textbf{Img.\ + Cap.} & \textbf{Image} & \textbf{Caption} & \textbf{Img.\ + Cap.} & \\ 
  \midrule
\multirow{4}{*}{\textbf{S}}
 & VisualBERT & 50.61 (+1.28) & 70.03 (-0.96) & 69.26 (-2.12) & 50.88 (+1.85) & 72.26 (-0.32) & 71.94 (-1.10) & 3\\
 & UNITER & 64.24 (\hlgreen{+14.09}) & 50.03 (\hlred{-20.72}) & 49.70 (\hlred{-21.47}) & 63.53 (\hlgreen{+13.55}) & 49.70 (\hlred{-19.81}) & 50.32 (\hlred{-20.45}) & 1\\
 & LXMERT & 33.27 (-0.24) & 33.27 (-0.24) & 33.27 (-0.24) & 33.43 (+0.38) & 33.43 (+0.38) & 33.43 (+0.38) & 0\\
 & ALBEF & 48.25 (-0.31) & 96.12 (+2.46) & 96.10 (+2.79) & 48.14 (+0.52) & 99.99 (-0.01) & 99.99 (-0.01) & 3\\
 \midrule
\multirow{4}{*}{\textbf{C}}
 & VisualBERT & 65.07 (\hlgreen{+19.87}) & 99.86 (-0.08) & 99.85 (-0.10) & 65.01 (\hlgreen{+18.39}) & 99.53 (-0.31) & 99.41 (-0.35) & 3\\
 & UNITER & 66.01 (\hlgreen{+23.65}) & 97.86 (-0.62) & 97.40 (-1.44) & 67.61 (\hlgreen{+23.28}) & 94.57 (-3.80) & 94.01 (-2.83) & 3\\
 & LXMERT & 60.43 (\hlgreen{+9.71}) & 54.03 (-0.97) & 54.35 (-0.91) & 64.42 (\hlgreen{+14.04}) & 76.80 (\hlgreen{+20.20}) & 73.63 (\hlgreen{+10.55}) & 3\\
 & ALBEF & 47.67 (+4.51) & 99.21 (-0.41) & 99.16 (-0.45) & 48.77 (+3.28) & 98.30 (-1.06) & 98.21 (-1.14) & 3\\
 \midrule
\multirow{4}{*}{\textbf{Q}}
 & VisualBERT & 66.34 (\hlgreen{+16.83}) & 99.01 (-0.97) & 99.01 (-0.96) & 65.35 (\hlgreen{+15.68}) & 98.43 (-1.56) & 98.37 (-1.63) & 3\\
 & UNITER & 75.04 (\hlgreen{+26.98}) & 93.41 (-4.44) & 93.82 (\hlred{-5.04}) & 73.18 (\hlgreen{+23.97}) & 88.14 (\hlred{-7.60}) & 88.43 (\hlred{-9.00}) & 3\\        
 & LXMERT & 64.72 (\hlgreen{+15.78}) & 53.23 (\hlgreen{+19.30}) & 50.13 (-0.90) & 62.36 (\hlgreen{+13.27}) & 69.16 (\hlgreen{+35.22}) & 67.06 (\hlgreen{+16.21}) & 3\\
 & ALBEF & 51.92 (+4.19) & 99.88 (-0.09) & 99.87 (-0.09) & 51.68 (+4.14) & 99.18 (-0.62) & 99.12 (-0.68) & 3\\
 \midrule
 \multirow{4}{*}{\textbf{N}}
 & VisualBERT & 59.32 (\hlgreen{+8.22}) & 92.68 (\hlred{-7.10}) & 92.79 (\hlred{-7.00}) & 61.54 (\hlgreen{+10.64}) & 99.62 (-0.27) & 99.64 (-0.26) & 3\\
 & UNITER & 61.79 (\hlgreen{+15.01}) & 56.92 (\hlred{-7.88}) & 56.85 (\hlred{-7.03}) & 76.65 (\hlgreen{+28.84}) & 99.17 (-0.34) & 99.08 (-0.71) & 3\\
 & LXMERT & 58.81 (\hlgreen{+8.13}) & 56.07 (-1.39) & 56.57 (-0.24) & 62.17 (\hlgreen{+13.80}) & 84.17 (-0.05) & 79.42 (\hlred{-19.24}) & 3\\
 & ALBEF & 42.80 (-4.44) & 88.82 (\hlred{-10.11}) & 88.83 (\hlred{-10.10}) & 47.34 (-1.40) & 99.62 (-0.22) & 99.62 (-0.22) & 3\\
 \midrule
  \bottomrule
\end{tabular}
\caption[Mean F$_1$ scores of models trained in the mixed setting.]{Mean F$_1$ scores of models trained in the mixed setting for the four V+L reasoning tasks (\textbf{S:} Spatiality, \textbf{C:} Cardinality, \textbf{Q:} Quantifiers, \textbf{N:} Numerical comparison). In parentheses, relative change from performance of models trained on image+caption setting are indicated. Differences larger than +5 are highlighted in \hlgreen{green} and differences smaller than -5 are highlighted in \hlred{orange}.} 
\label{tab:mixed1}
\end{center}
\vspace{-4mm}
\end{table*}

\FloatBarrier
\section{Modal dropout/``mixed'' setting}
\label{sec:mixed}

There is a concern that the bimodal presentation of the input in the image+caption setting facilitates overfitting to the textual modality. One way to alleviate the impact of bias towards either modality is by performing dropout of either the image or caption input in a subset of the input, thus introducing some unimodal examples into the image+caption dataset. This technique is inspired by the idea of sensor dropout as described in \citet{liu2017learning}, where overfitting to specific sensors (modalities) by a multisensor autonomous navigation system is avoided through dropout of features from sensors during training. In the following experiments, we perform dropout at a fixed probability of 25\% for both the image and caption modalities. In other words, datasets comprise 50\% image+caption, 25\% image-only and 25\% caption-only input. 

Overall, performance of models trained in the mixed setting resembles performance of models trained in the image+caption setting, in that they exhibit a bias towards the textual modality. Performance on the image modality is generally poorer than performance on both caption-only and image+caption test settings (with the exception of LXMERT), and performance on caption-only and image+caption settings are largely similar. Nevertheless, because of the dropout of the caption modality in 25\% of the input, the models are forced to also learn to deal with image-only input to some extent, and hence exhibit non-random performance on image-only test data in some cases.

\subsection{Comparison against models trained purely on image+caption input}

First, we compare the performance of models trained in the mixed setting against those trained in the image+caption setting. The relative difference is shown in brackets in Table \ref{tab:mixed1}. In the spatiality task, only UNITER achieves a non-random performance in the image-only setting, while VisualBERT, LXMERT and ALBEF exhibit no significant difference in comparison to the image+caption baseline. On the other tasks, we observe that VisualBERT, UNITER and LXMERT exhibit above random performance in the mixed setting, improving results over the purely image+caption setting. 

When models trained in the mixed setting are tested in the caption and image+caption settings, in the majority of cases, there is a small decrease in performance compared to models trained in the purely image+caption setting. The small magnitude of this difference is likely because the models continue to overfit to the caption when both image and caption input is presented, and are deprived of the textual information in only 25\% of the training data. The magnitude of the difference is larger in the numerical comparison task, where VisualBERT, UNITER and ALBEF exhibit significant decreases in performance on the OOD test set.

There are several exceptions to the overall trend. Firstly, unlike the other models, UNITER seems to have overfitted to the image modality instead of the caption modality in the spatiality task. Second, LXMERT exhibits an improvement in performance in both the caption and image+caption settings on the cardinality and quantifier datasets.

\subsection{Comparison against models trained on subset of data}

\begin{table*}[ht!]
\renewcommand{\arraystretch}{0.85}
\fontsize{10}{15}\selectfont
\begin{center}
\begin{tabular}{ @{} c|c|ccc|ccc|c @{} } 
%   \toprule
%   \multicolumn{9}{c}{\small \textbf{Spatiality}}\\
    \midrule
     \textbf{\uline{Task}} &  \textbf{\uline{Model}} 
   & \multicolumn{3}{c|}{\textbf{\uline{OOD}}} 
   & \multicolumn{3}{c|}{\textbf{\uline{InD}}} & \textbf{\uline{\#Seeds}} \\
   
& & \textbf{Image} & \textbf{Caption} & \textbf{Img.\ + Cap.} & \textbf{Image} & \textbf{Caption} & \textbf{Img.\ + Cap.} & \\ 
  \midrule
\multirow{4}{*}{\textbf{S}}
 & VisualBERT & +1.36 & \hlgreen{+19.07} & \hlred{-21.95} & +1.28 & \hlgreen{+21.66} & \hlred{-28.05} & 3\\
 & UNITER & \hlred{-27.24} & +0.37 & -0.66 & \hlred{-24.80} & -1.38 & +0.42 & 1\\
 & LXMERT & \hlred{-65.73} & \hlred{-16.15} & \hlred{-16.97} & \hlred{-65.59} & \hlred{-16.47} & \hlred{-17.65} & 0\\
 & ALBEF & +0.16 & \hlgreen{+47.09} & \hlgreen{+47.27} & -0.56 & \hlgreen{+50.87} & \hlgreen{+50.89} & 3\\
 \midrule
\multirow{4}{*}{\textbf{C}}
 & VisualBERT & \hlgreen{+6.49} & \hlgreen{+11.41} & +1.82 & +4.86 & \hlgreen{+12.50} & +4.48 & 3\\
 & UNITER & +1.52 & \hlgreen{+10.46} & +2.02 & +3.27 & \hlgreen{+11.35} & +4.31 & 3\\
 & LXMERT & -2.97 & +4.08 & +3.85 & -0.11 & \hlgreen{+19.47} & \hlgreen{+24.46} & 3\\
 & ALBEF & \hlred{-9.47} & \hlgreen{+12.29} & +0.47 & \hlred{-8.50} & \hlgreen{+9.64} & +1.83 & 3\\
 \midrule
 \multirow{4}{*}{\textbf{Q}}
 & VisualBERT & \hlgreen{+9.18} & \hlgreen{+44.72} & \hlgreen{+13.31} & \hlgreen{+8.70} & \hlgreen{+45.03} & \hlgreen{+15.75} & 3\\
 & UNITER & \hlgreen{+12.15} & \hlgreen{+43.48} & -2.30 & \hlgreen{+8.44} & \hlgreen{+36.99} & -2.73 & 3\\
 & LXMERT & \hlgreen{+31.32} & \hlgreen{+19.83} & +0.15 & \hlgreen{+28.89} & \hlgreen{+35.69} & \hlgreen{+17.93} & 3\\
 & ALBEF & +1.18 & \hlgreen{+7.37} & +2.14 & +0.81 & \hlgreen{+13.61} & +2.92 & 3\\
  \midrule
\multirow{4}{*}{\textbf{N}}
 & VisualBERT & \hlgreen{+6.73} & \hlgreen{+38.46} & \hlgreen{+17.75} & +0.09 & \hlgreen{+43.60} & +1.18 & 3\\       
 & UNITER & +0.10 & \hlred{-8.21} & -3.95 & +4.69 & \hlgreen{+5.98} & +0.26 & 3\\
 & LXMERT & -0.59 & -2.77 & -1.28 & +1.54 & \hlgreen{+18.20} & \hlgreen{+11.76} & 3\\
 & ALBEF & -2.75 & \hlred{-5.73} & \hlgreen{+28.78} & -0.61 & \hlgreen{+5.29} & \hlgreen{+21.15} & 3\\
  \bottomrule
\end{tabular}
\caption[Difference in F$_1$ scores between models trained on the mixed setting and models trained on subset baseline.]{Difference in F$_1$ scores between models trained on the mixed setting and models trained on subset baseline for the four V+L reasoning tasks (\textbf{S:} Spatiality, \textbf{C:} Cardinality, \textbf{Q:} Quantifiers, \textbf{N:} Numerical comparison). Differences larger than +5 are highlighted in \hlgreen{green} and differences smaller than -5 are highlighted in \hlred{orange}.} 
\label{tab:mixed3}
\end{center}
\vspace{-4mm}
\end{table*}
\FloatBarrier

Next, we compare the results from the mixed setting against a baseline if the mixed dataset were separated into datasets with fully unimodal or bimodal examples. For instance, the 32k mixed dataset on the spatiality task can be compared against baseline datasets of i) 16k bimodal image+caption examples, ii) 8k image-only examples and iii) 8k caption-only examples. Specifically, we ask if there is a performance benefit to training models on a single mixed dataset which combine these subsets together, as opposed to training models on each dataset separately.

This baseline is obtained from training on a randomly sampled subset of the full dataset. As previously explained, these experiments on subsets of the full dataset were conducted only on a single random seed, due to resource limitations. In Table \ref{tab:mixed3}, we observe that the performance of models trained on the mixed dataset indeed outperforms models trained on only the caption-only and image+caption subset of the data. This result is not unexpected, given that it is known that models exploit the caption present in the image+caption examples.

However, it is less clear from the results whether the models are able to exploit the additional image information present in the image+caption examples in the mixed setting to improve performance on the image-only test examples. In particular, results from the quantifier task, as well as the results of VisualBERT on the OOD test set seem suggestive of some advantage accorded to models by the additional image+caption data. Nevertheless, there are also results suggestive of the opposite finding, particularly on the spatiality task, where the presence of caption-only and image+caption examples degrades performance in the image-only setting.

We conclude that the technique of dropping out the input from some modalities helps to ameliorate the extent of the textual bias in the image+caption setting to some extent, although the overall textual bias of models remains.

\section{Error analysis}
\label{sec:erroranalysisapp}
In section \ref{sec:erroranalysis}, we described our error analysis of the spatiality and cardinality tasks. In this appendix section, we include additional details on the cardinality task and present our error analysis on the quantifiers and numerical comparison tasks.

\subsection{Cardinality}

\begin{figure}[h!]
\small
    \centering
    \includegraphics[width=0.9\linewidth]{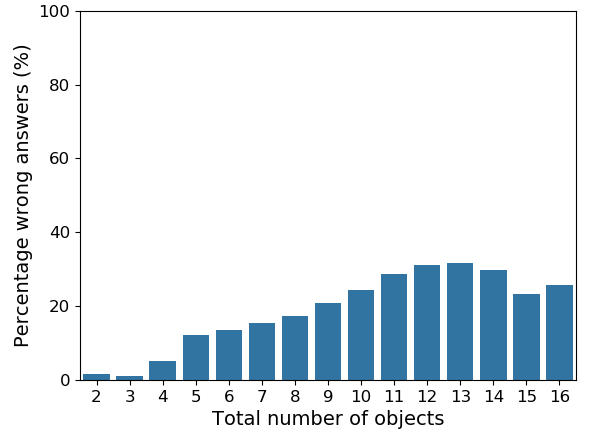}
    \caption[Percentage of incorrect answers by total number of objects.]{Percentage of incorrect answers by total number of objects. VisualBERT trained on cardinality task, image-only setting, tested on OOD test set. (F$_1$=77.84; seed=0)}
    \label{fig:analysis_card}
\end{figure}

As noted in section \ref{sec:erroranalysis}, a significant factor is the total number of objects in the image, inclusive of distractors. Figure~\ref{fig:analysis_card} shows an increasing percentage of wrong answers as the total number of objects increases. We also notice a slight decrease in wrong answers when the number of objects exceeds 13. This can be attributed to the decreasing likelihood of correct answers having smaller numbers.

\subsection{Quantifiers}

\begin{figure}[h!]
\small
\centering
\begin{subfigure}{0.23\textwidth}
\includegraphics[width=\textwidth]{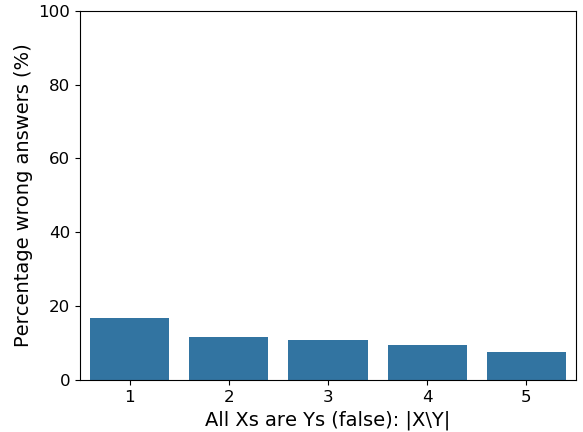}
\caption{All Xs are Ys (false): $|X\setminus Y|$}
\end{subfigure}
\hspace{\fill}
\begin{subfigure}{0.23\textwidth}
\vspace{0.35em}
\includegraphics[width=\textwidth]{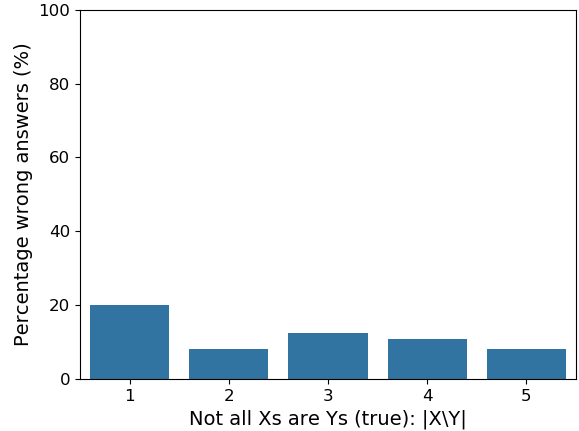}
\caption{Not all Xs are Ys (true): $|X\setminus Y|$}
\end{subfigure}
\hspace{\fill}
\begin{subfigure}{0.23\textwidth}
\vspace{0.35em}
\includegraphics[width=\textwidth]{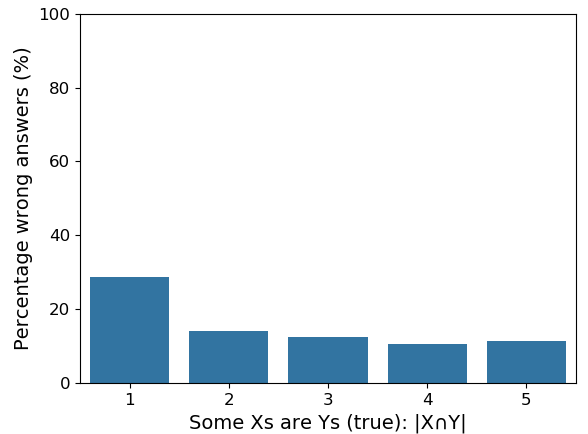}
\caption{Some Xs are Ys (true): $|X\cap Y|$}
\end{subfigure}
\hspace{\fill}
\begin{subfigure}{0.23\textwidth}
\includegraphics[width=\textwidth]{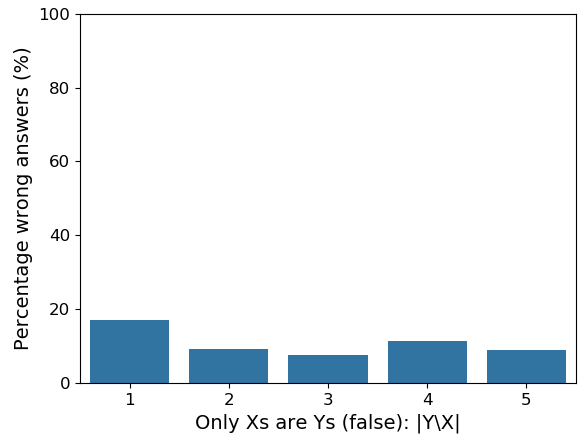}
\caption{Only Xs are Ys (false): $|Y\setminus X|$}
\end{subfigure}
\hspace{\fill}
\begin{subfigure}{0.23\textwidth}
\vspace{0.35em}
\includegraphics[width=\textwidth]{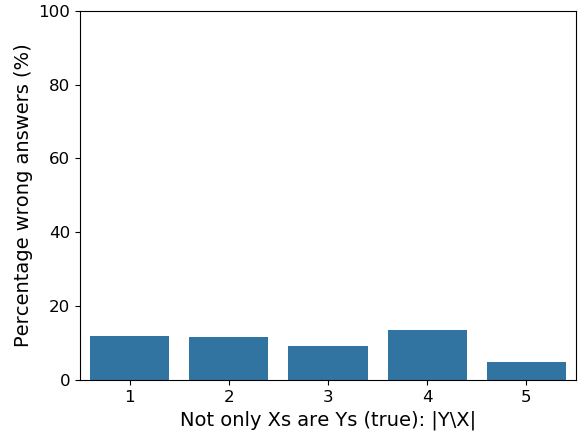}
\caption{Not only Xs are Ys (true): $|Y\setminus X|$}
\end{subfigure}
\hspace{\fill}
\begin{subfigure}{0.23\textwidth}
\vspace{0.35em}
\includegraphics[width=\textwidth]{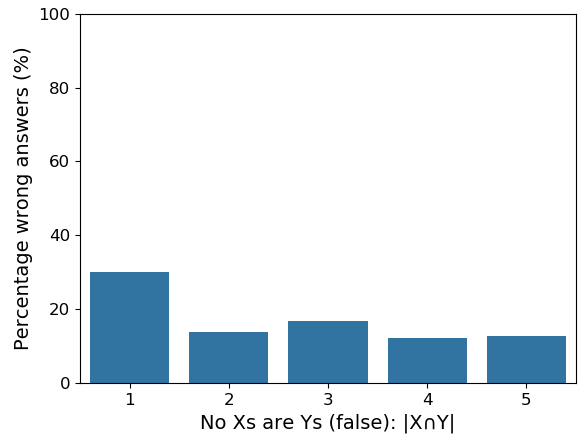}
\caption{No Xs are Ys (false): $|X\cap Y|$}
\end{subfigure}
\caption[Percentage of wrong answers on quantifier-specific values.]{Percentage of wrong answers on quantifier-specific values. VisualBERT, image-only setting on InD test set. (F$_1$=87.98; seed=0)}
\label{fig:analysis-quant}
\end{figure} 

Figure~\ref{fig:analysis-quant} plots the percentage of wrong answers against specific values relevant to the particular quantifiers of VisualBERT on the InD test set in the image-only setting. Similar patterns are also observed across other models in runs with moderate performance, although moderate performance in some cases may be due to failure to learn a subset of quantifiers. For instance, Figure~\ref{fig:analysis-quant}a shows the percentage of wrong answers against the number of Xs which are not Ys, given all \textit{false} queries of the form ``All Xs are Ys''. The results show that the more Xs which are not Ys, the less likely it is for the model to incorrectly predict the query to be true. In Figure~\ref{fig:analysis-quant}c, given all \textit{true} queries of the form ``Some Xs are Ys'', the more objects which are both X and Y, the less likely it is for the model to incorrectly predict the query to be false. Although the trend is not always robust across all the quantifiers, this overall pattern of errors made by models is not unexpected.

\subsection{Numerical comparison}

\begin{figure*}[h!]
\small
\centering
\begin{subfigure}{0.45\textwidth}
\includegraphics[width=\textwidth]{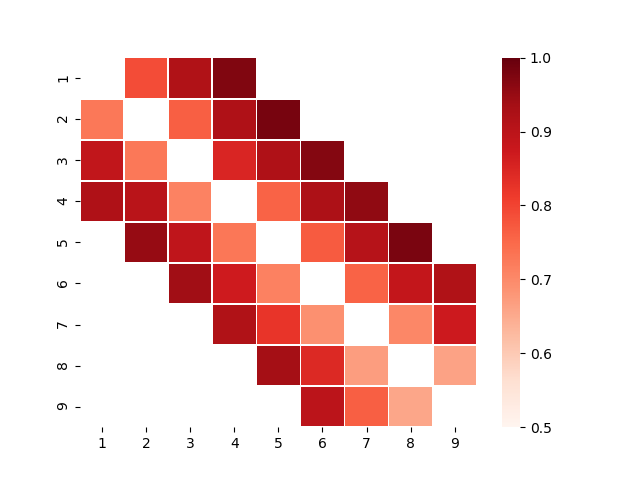}
\caption{LXMERT, image-only InD. (F$_1$=84.45; seed=0)}
\end{subfigure}
\hspace{\fill}
\begin{subfigure}{0.45\textwidth}
\vspace{0.25em}
\includegraphics[width=\textwidth]{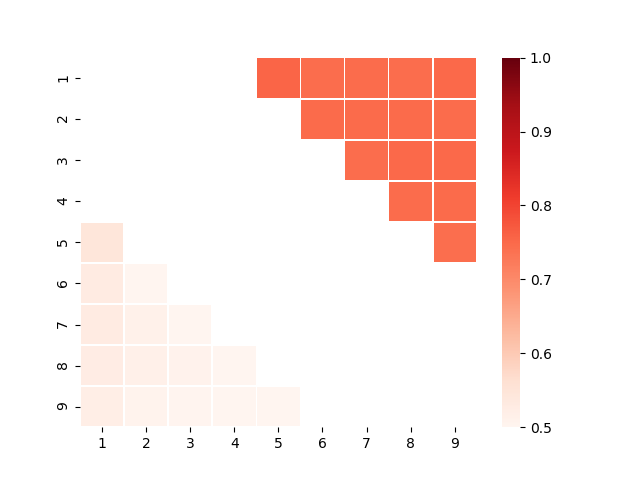}
\caption{LXMERT, image-only OOD. (F$_1$=63.01; seed=0)}
\end{subfigure}
\caption{Results of LXMERT, image-only setting on specific $<x, y>$ pairs.}
\label{fig:analysis-comp1}
% \end{figure} 

% \begin{figure}[!htb]
\small
\centering
\begin{subfigure}{0.40\textwidth}
\includegraphics[width=\textwidth,valign=t]{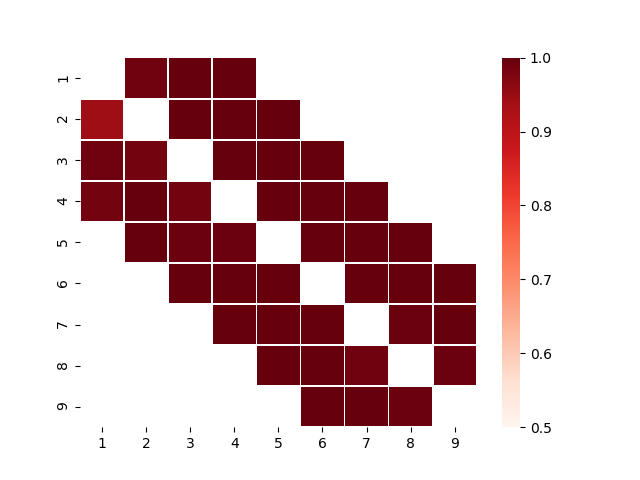}
\caption{UNITER, caption-only InD. (F$_1$=99.66; seed=0)}
\end{subfigure}
\hspace{\fill}
\begin{subfigure}{0.40\textwidth}
\vspace{0.25em}
\includegraphics[width=\textwidth,valign=t]{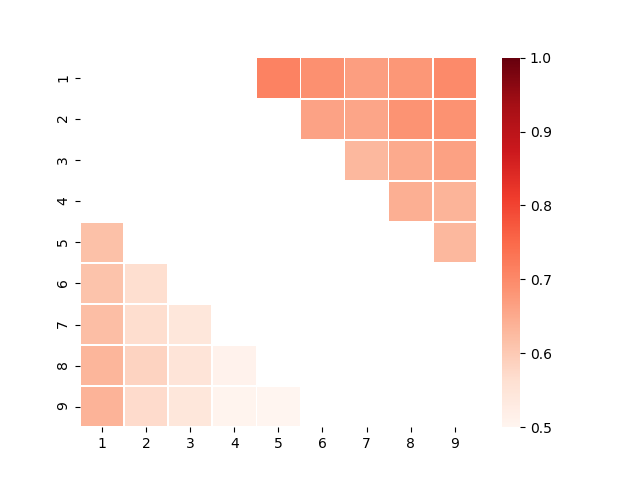}
\caption{UNITER, caption-only OOD. (F$_1$=61.90; seed=0)}
\end{subfigure}
\caption{Results of UNITER, caption-only setting on specific $<x, y>$ pairs.}
\label{fig:analysis-comp2}
\end{figure*}

The plot in Figure~\ref{fig:analysis-comp1} reveals a clear trend where models are more accurate on pairs where the magnitude of $|x-y|$ (i.e. the difference between the numerals being compared) is larger for the InD test set. However, the more significant finding is that the numerical comparison task was the only task which exhibited significant differences in performance between the InD and OOD datasets, due to an inability to generalise to unseen pairs particularly in the image-only setting, but also the caption-only setting in the case of UNITER and LXMERT.

In Figure~\ref{fig:analysis-comp1}, we observe a pattern where the model succeeds only on OOD queries where the number of the first object exceeds that of the second object (cells on the upper right are darker than cells on the lower left). This asymmetry is also observed to some extent in Figure~\ref{fig:analysis-comp2}. This pattern does not result from any simple strategy of providing the same answer (true/false) given a relation in the query (\textit{more}/\textit{fewer}). A similar pattern is observed even when subsetting the OOD test set into queries with either the \textit{more} or \textit{fewer} relation in both cases. The overall finding is that models are able to generalise to unseen number pairs by constructing an implicit numeral scale, but only to a limited extent.

\end{document}